\documentclass[conference]{IEEEtran}

\usepackage[utf8]{inputenc}
\usepackage[T1]{fontenc}
\usepackage{amsmath,amssymb,amsfonts,amsthm,mathtools}
\usepackage{bm}
\usepackage{hyperref}
\usepackage[dvipsnames]{xcolor}

\usepackage[numbers]{natbib} %for citations
\usepackage{tikz}
\usetikzlibrary{positioning,arrows.meta,calc}
\usepackage{float}

\definecolor{solutioncolor}{HTML}{0072B2}
\definecolor{correctioncolor}{HTML}{D55E00}
\definecolor{neutral}{HTML}{323943}

\usepackage{pgfplots}
\pgfplotsset{compat=1.18}

\newcommand{\R}{\mathbb{R}}
\newcommand{\eps}{\varepsilon}

\newcommand{\pinv}{^{\dagger}}
\newcommand{\norm}[1]{\left\lVert #1 \right\rVert}
\newcommand{\im}{\operatorname{Im}}
\newcommand{\Id}{\mathrm{Id}}

\newcommand{\bmu}{\bm{\mu}}

\definecolor{viridisteal}{RGB}{62, 72, 136}

\definecolor{linkblue}{RGB}{0,70,160}

\makeatletter
\renewcommand{\subsubsection}{%
  \@startsection{subsubsection}{3}{\parindent}%
  {0.8ex plus 0.2ex minus 0.1ex}% espace avant le titre
  {0.3ex}% espace après le titre, avec retour à la ligne
  {\normalfont\normalsize\itshape}%
}
\makeatother

\title{Learning features from Newton's algorithm: a way to accelerate nonlinear parametrized PDE solvers}

\renewcommand{\IEEEauthorrefmark}[1]{%
  \raisebox{0pt}[0pt][0pt]{%
    \textsuperscript{\footnotesize #1}%
  }%
}

\author{
\IEEEauthorblockN{
Rémy Vallot\IEEEauthorrefmark{1,2},
Florian De Vuyst\IEEEauthorrefmark{3},
Thibault Dairay\IEEEauthorrefmark{1,2},
Mathilde Mougeot\IEEEauthorrefmark{1,4}
}

\vspace{0.8em}

\IEEEauthorblockA{
\IEEEauthorrefmark{1}
Centre Borelli UMR 9010, ENS Paris-Saclay,
Université Paris-Saclay, CNRS, France
}

\IEEEauthorblockA{
\IEEEauthorrefmark{2}
Michelin, Centre de Recherche de Ladoux,
Cébazat, France
}

\IEEEauthorblockA{
\IEEEauthorrefmark{3}
BMBI UMR 7338, Université de Technologie de Compiègne, CNRS, France
}

\IEEEauthorblockA{
\IEEEauthorrefmark{4}
ensIIE, Evry, France
}
}

\begin{document}

\maketitle
\thispagestyle{plain} 
\pagestyle{plain}   
\begin{abstract}

~It is well known that Newton's method converges faster when the initial guess is closer to a root of a system of nonlinear equations.

In this paper, a two-stage Newton initial guess strategy is proposed by learning features from a parameter-space sampling and a database of precomputed solutions. 
The method uses discrete Newton trajectories to construct two complementary reduced spaces: a solution feature space, built from converged states, and a corrective search direction feature space, built from intermediate Newton increments.
For an unseen parameter, a regression model is used to predict a surrogate solution approximation. Then, in a second step, a residual-minimizing correction is computed using a dedicated GMRES-based approach.
The resulting state is then used as an initial guess for the high-fidelity Newton method, which completes convergence.

The corrective step is computationally inexpensive since it only requires residual evaluations and the solution of a small least-squares problem.  The methodology is weakly intrusive once the high-fidelity residual fields and a script-based programming interface are available.

This strategy reduces the number of Newton iterations and decreases the overall CPU time. Numerical experiments on representative PDE problems show quantifiable speedups compared with standalone surrogate initialization.
Significant speedups are observed. This generic approach can be applied to a broad class of large-scale nonlinear problems.

\end{abstract}


% --------------------------------------------------
% SECTION I: INTRODUCTION
% --------------------------------------------------

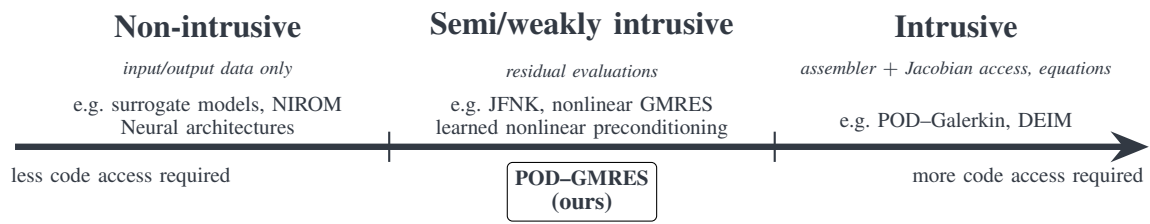
\begin{figure*}[t]
\centering
\begin{tikzpicture}[
  head/.style={font=\bfseries\large},
  acc/.style={font=\scriptsize\itshape},
  ex/.style={font=\footnotesize},
  tick/.style={thick, draw=neutral,},
  text=neutral,
]

\coordinate (Astart) at (0,0);
\coordinate (Aend)   at (15,0);

\coordinate (Cni) at ($(Astart)!0.17!(Aend)$);
\coordinate (Csi) at ($(Astart)!0.50!(Aend)$);
\coordinate (Cin) at ($(Astart)!0.83!(Aend)$);

\coordinate (T1) at ($(Astart)!0.33!(Aend)$);
\coordinate (T2) at ($(Astart)!0.67!(Aend)$);

% category labels above the arrow
\node[head] (h1) at ($(Cni)+(0,1.6)$) {Non-intrusive};
\node[acc, below=0.8mm of h1] (s1) {input/output data only};
\node[ex, below=0mm of s1] (e1) {e.g. surrogate models, NIROM};
\node[ex, below=3mm of s1] (e1) {Neural architectures};

\node[head] (h2) at ($(Csi)+(0,1.6)$) {Semi/weakly intrusive};
\node[acc, below=0.8mm of h2] (s2) {residual evaluations};
\node[ex, below=0mm of s2] (e2) {e.g. JFNK, nonlinear GMRES};
\node[ex, below=3mm of s2] (e2) {learned nonlinear preconditioning};

\node[head] (h3) at ($(Cin)+(0,1.6)$) {Intrusive};
\node[acc, below=0.8mm of h3] (s3) {assembler $+$ Jacobian access, equations};
\node[ex, below=2mm of s3] (e3) {e.g. POD--Galerkin, DEIM};

% the thick backbone arrow
\draw[draw=neutral, -{Stealth[length=5mm,width=4mm]}, line width=2.4pt] (Astart) -- (Aend);
\draw[tick] ($(T1)+(0,-2mm)$) -- ($(T1)+(0,2mm)$);
\draw[tick] ($(T2)+(0,-2mm)$) -- ($(T2)+(0,2mm)$);
\node[font=\footnotesize] at ($(Csi)+(5.9,-0.4)$) {more code access required};
\node[font=\footnotesize] at ($(Csi)+(-6.1,-0.4)$) {less code access required};

% marker for "this work", placed within the semi-intrusive segment
\coordinate (Mine) at ($(Astart)!0.5!(Aend)$);
\node[align=center, font=\bfseries\footnotesize, fill=white, draw, rounded corners=2pt,
      inner sep=3pt, below=2mm of Mine] (mlabel) {POD--GMRES\\\small(ours)};

\end{tikzpicture}
\caption{Positioning of the proposed method along the intrusiveness spectrum. Non-intrusive methods use only input/output data; intrusive methods require access to the assembler and Jacobian; the proposed POD--GMRES lies in the semi-intrusive middle ground, requiring only residual evaluations.}
\label{fig:positioning}
\end{figure*}

% =====================================================================
\section{Introduction}

The numerical discretization of nonlinear partial differential problems typically results in the solution of a large-scale nonlinear algebraic system. The problem is formalized as follows:
\begin{equation}
\bm{F}(\bm{u},\bm{\mu})=\bm{0}, \qquad \bm{F}:\R^{n}\times\mathcal{P}\to\R^{n},
\label{eq:nonlinear-pb}
\end{equation}
where $\bm{u}\in\R^{n}$ is the discrete solution, $\bm{\mu}\in\mathcal{P}$ the parameter vector living in a bounded domain $\mathcal{P}$ of $\mathbb{R}^p$, and $\bm{F}$ the nonlinear residual operator.

To solve such problems, Newton and quasi-Newton methods, possibly combined with Krylov-subspace techniques (Generalized Minimal Residual, Jacobian-Free Newton--Krylov approaches), remain standard computational tools~\cite{saad2001iterative,vandervorst2003krylov,Knoll_2004}.

Solving~\eqref{eq:nonlinear-pb} with a full-order Newton method at each parameter query is a computationally expensive task.
The stronger the nonlinearity of the equation, the more iterations may be required.
Each iteration involves assembling a Jacobian matrix of size $n\times n$ and solving a large-scale linear system. Most of the time, iterative methods themselves are used for that task (preconditioned conjugate gradient method, etc).
The residual vector must also be assembled in every iteration, but its cost can be considered negligible compared to that of the Jacobian matrix.
%.
This global cost is further multiplied in many-query contexts such as design, optimization, and uncertainty quantification, in which~\eqref{eq:nonlinear-pb} must be solved for many parameter values. This has historically motivated model-order reduction. Repeated Newton solves are expensive, but they also generate reusable information.
In this work, Newton's algorithm is regarded not only as a nonlinear solver but also as a source of features that can be learned and reused. Two complementary feature spaces are extracted from the offline computations: a solution space learned from the converged states and a correction space learned from the intermediate Newton increments.

The efficiency of Newton-type solvers is highly sensitive to the quality of initialization.
In the absence of a priori knowledge or available precomputed solutions, one typically uses a simple field satisfying the boundary conditions, but this provides no convergence guarantee.
For strongly nonlinear problems, a continuation method is often used to correctly initialize the fixed-point method, but the price to pay is a parameter continuation outer loop, leading to a very expensive procedure \cite{numerical-continuation}.

When precomputed solutions in the admissible parameter space are available, a natural strategy is first to build a surrogate model to infer a proper initialization. 
Depending on the amount of available data, different initialization strategies can be used. A minimalist approach is to initialize with the solution of the nearest-neighbor parameter sample. 
More sophisticated approaches use dimensionality reduction and model-order reduction (reduced-basis methods, proper orthogonal decomposition, empirical interpolation method, approximate tensor decomposition, etc.) in a limited-data regime. If data scarcity is not a problem, one can imagine using more highly parametrized nonlinear architectures.

The solution space alone, however, limits the acceleration that can be achieved by surrogate-based initialization. 
For example, using a proper orthogonal decomposition (considering a Hilbert space $X$ with its scalar product $(.,.)_X$) added with some POD-coefficient regression model,
the surrogate solution will be in the form of the approximate tensor decomposition
\[
\hat u(.,\bm{\mu}) = \sum_{k=1}^K a_k(\bm{\mu}) \, \psi_k(.)
\]
where the $\psi_k$ are the POD modes, $a_k(\bm{\mu})\approx (u(.,\bmu),\psi_k)_X$ are the approximate POD coefficients and $K$ is the low-order POD truncation rank. The solution error $u(.,\bmu)-\hat u(.,\bmu)$ depends on both the truncation rank and the POD-coefficients regression model in use.

One could try to improve the linear-combination coefficients through an inner iterative process. However, as long as the approximation remains within the fixed POD space, its error cannot be reduced below the orthogonal projection error. Consequently, optimizing the coefficients alone cannot overcome the representational limitation of the solution feature space and may lead to a residual plateau.
This approach allows starting the Newton method with a reasonably accurate initial guess, but possibly not sufficiently close to the solution to achieve superlinear convergence, limiting the global speedup.

This limitation motivates the need to extend the search space. A machine-learning algorithm is proposed to return search direction features.
A second step residual-based procedure allows one to get additional suitable search directions. Then an inexpensive least-squares minimization problem is solved, returning a better approximate solution.

Finally, a few Newton iterations are done until the stopping accuracy criterion is reached. This last step is required to measure the speedup factors with comparable accuracy.
\medskip

The key features of this proposed methodology are the following:
\begin{itemize}
    \item Two complementary feature spaces are learned from previous offline Newton computations: a solution space constructed from the converged states and used for prediction, and a corrective search direction space constructed from the intermediate Newton increments and used for residual reduction.
    \item  A weakly-intrusive Jacobian-free correction step in the spirit of GMRES is proposed. A linearized residual least-squares problem is minimized over a low-order offline corrective subspace of dimension $r$. Each step costs $(r+1)$ residual evaluations and a least-squares problem solution, making it computationally weakly expensive.
    \item The correction drives the residual norm below the POD-based regression-error plateau of the surrogate initialization, which translates, through the iteration bound of \cite{jin2024}, into fewer expensive Newton iterations.
\end{itemize}

The paper is organized as follows. Section~\ref{sec:related} gives an overview of similar and related works. Section~\ref{sec:methodology} presents the proposed method in detail. Section~\ref{sec:experiences} describes the case studies for numerical experiments. The results, followed by a discussion, are presented in Section~\ref{sec:results}. Finally, the paper concludes with remarks and perspectives in Section~\ref{sec:conclusion}.

\section{Related work} \label{sec:related}
%====================

To reduce the time to solution, various strategies can be applied. These strategies are classified according to their level of intrusiveness, i.e., the degree of access required to the solver. An overview is given in Fig.~\ref{fig:positioning}.
At one end, non-intrusive methods rely solely on input–output data from the solver. In this case, the solver is treated as a black box that generates the data needed to construct surrogate models.
At the opposite end, intrusive methods require access to the assembler, the Jacobian matrices, and the ability to modify the solver (reduced residuals, etc.).
Between these two, semi-intrusive methods require some internal knowledge of the solver, such as the evaluation of the residual operator, but are still less intrusive than approaches that need full access to the assembler or the discretization scheme.

This section reviews the acceleration literature through that lens and situates the proposed methodology within it.

Throughout, the comparison follows a fair comparison protocol: every comparison is made at the same accuracy as the converged high-fidelity solution. This is consistent with the protocol defined by \citet{weakbaselines} for evaluating speedups. 
 
\subsection{Non-intrusive acceleration}
 
The least intrusive way to accelerate a nonlinear solve is to improve its starting point.

The system~\eqref{eq:nonlinear-pb} is assumed to have already been solved for a set of training parameters
$\{\bm{\mu}_{1},\dots,\bm{\mu}_{p}\}\subset\mathcal{P}$, yielding the
converged high-fidelity solutions $\{\bm{u}^{\star}(\bm{\mu}^{(i)})\}_{i=1}^{p}$.

A surrogate model $\hat{\bm{u}}:\mathcal{P}\to\R^{n}$ is then built from these parameter-solution pairs, mapping a parameter to an approximate solution $\hat{\bm{u}}$. 
Hence, for an unseen parameter $\bm{\mu}$, the prediction $\hat{\bm{u}}(\bm{\mu})$ replaces the constant field as the Newton initial iterate $\bm{u}^{(0)}$ in~\eqref{eq:newton}.

Since $\hat{\bm{u}}(\bm{\mu})$ generally lies closer to the solution, the solver needs fewer iterations, which already reduces the cost of a single query.

This approach is theoretically supported by the analysis of \citet{jin2024}, which establishes an upper bound on the number of Newton iterations in terms of the initial solution error under local regularity assumptions.
If the initial error is reduced by a factor $2^{-n}$, this bound decreases logarithmically in $n$. 
Although the residual norm is not the quantity appearing in this bound, it is used here as an empirical indicator of initialization quality.
A surrogate that lowers the initial residual therefore reduces the number of expensive iterations.

Such surrogates are inexpensive to evaluate and require only input-output data. This is a non-intrusive approach, and different classes of surrogate models, such as reduced-order models ~\cite{podgp, quadraticmanifolds}, neural networks and neural operators~\cite{deeponet,FNO,GKN,raonietal2023} all belong to this category.
 
Surrogate-based initialization has been used across several domains, including solid mechanics~\cite{taghikhanietal2025,Kadeethum_2022}, fluid-structure interaction~\cite{tiba2022fsi,bergmann2016ot}, stiff time-dependent PDEs under implicit time stepping~\cite{jin2024}, nonlinear elliptic PDEs~\cite{FNO_efranck}, subsurface flow~\cite{lechevallier}, power systems~\cite{donti2020smartpgsim}, reacting and hypersonic flows~\cite{novello2022nn}, and general nonlinear algebraic systems~\cite{intdeep}.

Most of these contributions are empirical and do not provide theoretical guarantees. In contrast, \citet{jin2024} established an upper bound on the number of Newton iterations required to reach convergence.

The surrogate evaluation introduces no additional physical information before the high-fidelity solver takes over, so the attainable initial residual $\lVert\bm{F}(\hat{\bm{u}},\bm{\mu})\rVert$ is bounded below by the surrogate's regression error. As a result, the surrogate alone cannot deliver the high-fidelity solution. This leads to a common issue found in the literature, where some replace the solver output with the surrogate prediction and compare the time to solution directly for speedup determination. This is misleading because the comparison is no longer made at fixed accuracy. As mentioned before, the surrogate prediction is only guaranteed to be accurate up to its regression error, which is too large to satisfy the convergence criterion. 
Therefore, it must be used only as an initialization for the high-fidelity solve rather than as a full substitute.

\subsection{Intrusive acceleration}
Intrusive methods, as their name indicates, require deeper access to the solver but can achieve larger speedups. Several types of intrusive methods exist, including projection-based and preconditioning approaches. Projection-based reduced-order models project the governing equations onto a low-dimensional subspace. For example, POD-Galerkin projection, least-squares Petrov-Galerkin, and the GNAT method~\cite{GNAT} reduce the dimension of the system. Hyper-reduction methods based on empirical interpolation make it possible to evaluate the nonlinearity at only a small set of indices~\cite{EIM,DEIM,gappyPOD}. 
These techniques require access to the assembler and the Jacobian, as well as the ability to implement solver logic on reduced coordinates.

Preconditioning transforms the system $\bm{F}(\bm{u})=\bm{0}$ into an equivalent system with the same solution but reduced nonlinearity, so that Newton's method converges faster or more easily. Methods such as ASPIN~\citep{caikeyes2002aspin} and RASPEN~\citep{dolean2016raspen} fall in this domain, both built from an overlapping domain decomposition with local nonlinear subproblem solves. 
At the linear level, the Jacobian system solved at each Newton step is itself accelerated by classical preconditioning techniques~\cite{benzi2002preconditioning,chen2005preconditioning,axelsson1996iterative}.

For both hyper-reduction and preconditioning techniques, intrusiveness is often an obstacle in an industrial setting, where the solver is frequently a black box exposing little to no internal access. These methods are also highly problem-dependent. This restriction motivates the search for acceleration techniques that require less access to the internal structure of the solver.

\subsection{Semi-intrusive acceleration}
 
Between the two strategies lies a category of methods that requires more than input-output solution data but not full access to the discretization machinery or the assembler. These methods require only the ability to evaluate the residual operator $\bm{F}(\bm{u},\bm{\mu})$ without ever assembling or inverting the Jacobian and without modifying the discretization.
Calling the residual operator is made possible by most industrial solvers, while modifying the discretization may be difficult and needs deeper access to the code. 

Some surrogate-based initialization strategies can be considered semi-intrusive because they require governing equations, residuals, or weak forms to regularize the learning process. Such methods fall into physics-informed neural architectures (networks and operators), equation-aware neural operators \cite{wang2021learningsolutionoperatorparametric, no_nvidia, li2023physicsinformedneuraloperatorlearning, cheng2026}. \citet{cheng2026} used such a surrogate to initialize a PDE solver and ensure the same accuracy comparison.

However, incorporating physical knowledge into neural architectures comes at a non-negligible offline cost, as the optimization problem can be computationally expensive and making the optimization converge can be tricky. This additional cost must also be amortized over a sufficiently large number of online queries. Otherwise, a simpler surrogate that is less accurate but substantially cheaper to construct may provide a better overall time-to-solution.

\citet{lee2025neural} propose a neural-operator preconditioned Newton in which the neural operator learns to map the current iterate toward the solution.

\citet{dingwang2025} propose a residual-driven adaptive Newton strategy. The residual reweighting balances unbalanced nonlinearities and helps the Newton-type solver select a more appropriate step length~\cite{dingwang2025}.

\citet{pinl} propose to use Proper Orthogonal Decomposition to learn components of the inexact-Newton residual from training problems, allowing the definition of a projected low-dimensional Jacobian system solved using Newton. The converged solution of the projected system is then used as an initialization for the inexact Newton method.

A large part of the semi-intrusive acceleration literature focuses on the linear systems solved inside Newton's method rather than on the nonlinear Newton iteration itself. 
At each Newton step, the correction is obtained by solving a Jacobian system, and Krylov methods such as GMRES accelerate this linear solve without explicitly inverting the Jacobian~\cite{saad2001iterative,vandervorst2003krylov}. Classical preconditioning acts at the same level by improving the conditioning of this linearized system~\cite{benzi2002preconditioning,chen2005preconditioning,axelsson1996iterative}. These approaches reduce the cost of the inner linear solve, but they do not directly modify the nonlinear residual map $\bm{F}$ or the Newton initialization. This distinction is important here, since the proposed correction uses a GMRES-type residual minimization at the nonlinear level, before the high-fidelity Newton solver is called.

\subsection{Positioning of the proposed method}

Existing surrogate-initialization approaches primarily learn solution features or a direct parameter-to-solution map from converged high-fidelity states. They generally do not exploit the intermediate trajectories generated during the offline nonlinear solves.

The proposed method combines a non-intrusive prediction stage with a semi-intrusive residual-based correction stage. This work is set in the semi-intrusive configuration (Fig.~\ref{fig:positioning}).

This distinction separates the proposed approach from methods that learn only from solution features and from intrusive reduced-order methods that replace the high-fidelity nonlinear system.

Since it does not provide global convergence guarantees, the correction is not intended to replace the high-fidelity Newton solver. Instead, it drives the residual down with Jacobian-free steps built on an offline basis of corrective features and then hands the improved iterate to the full-dimensional Newton solver.

The correction offers no guarantee of convergence to the global minimum of the residual. No such guarantee is claimed, and in practice, the iteration may stall at a nonzero residual. Nonetheless, this is not a limitation for the intended purpose. Any reduction of the residual below the surrogate's regression plateau has a direct impact on the iteration-count bound of~\cite{jin2024}, resulting in fewer expensive Newton iterations remaining. This has a strong impact on the total computation time to obtain the solution.

The high-fidelity solver is kept and still delivers the converged solution, so the equal-accuracy requirement is met by construction. The enhancement strategy only improves surrogate initialization.
In this sense, the method is strictly less intrusive than POD-Galerkin reduction or ASPIN-type nonlinear preconditioning, as it requires no Jacobian call, no domain decomposition, and no modification of the solver while still reintroducing some physical knowledge that a pure predictor gives up.

% SECTION : METHODOLOGY & POD INITIALIZATION
% --------------------------------------------------
\section{Methodology}
\label{sec:methodology}

\begin{figure*}[!t]
\centering

\resizebox{\textwidth}{!}{%

\begin{tikzpicture}[
  text=neutral,
  head/.style={font=\bfseries\large},
  acc/.style={font=\footnotesize\itshape,align=center},
  ex/.style={font=\footnotesize,align=center},
  legend/.style={font=\footnotesize,anchor=west},
  box/.style={draw=neutral,thick,rounded corners=2pt,fill=white,font=\footnotesize,
    align=center,inner xsep=4pt,inner ysep=4pt},
  flow/.style={-{Stealth[length=3mm,width=2.3mm]},line width=1.15pt,draw=neutral},
  blueflow/.style={flow,draw=solutioncolor},
  orangeflow/.style={flow,draw=correctioncolor},
  neutralflow/.style={flow,draw=neutral},
  divider/.style={draw=neutral,line width=1.1pt}
]

% ---------------------------------------------------------------------------
% OFFLINE: learn two spaces from complete Newton trajectories
% ---------------------------------------------------------------------------
\node[head,anchor=west] (offline) at (-0.7,5.45)
  {\textcolor{neutral}{Offline feature learning}};
\node[acc,anchor=west] at (-0.7,5.0)
  {Learning features from Newton's full trajectories generated for the training problems};

% Color legend
\draw[draw=solutioncolor,line width=2.2pt] (7.15-3,-3.2) -- (7.60-3,-3.2);
\node[legend,text=neutral] at (7.72-3,-3.2) {solution features};
\draw[draw=correctioncolor,line width=2.2pt] (11.05-4,-3.2) -- (11.50-4,-3.2);
\node[legend,text=neutral] at (11.62-4,-3.2) {corrective features};
\draw[draw=neutral,line width=2.2pt] (14.55-4.5,-3.2) -- (15.00-4.5,-3.2);
\node[legend,text=neutral] at (15.12-4.5,-3.2) {high-fidelity solve};

\node[box,text width=2.4cm] (queries) at (0.7+1.2,3.55-0.3) {
  \textbf{Training queries}\\[1mm]
  $\{\boldsymbol\mu_i\}_{i=1}^{N_{\mathrm{train}}}$
};

\node[box,text width=3.4cm] (trajectories) at (4.5+1.2,3.55-0.3) {
  \textbf{High-fidelity Newton}\\[0mm]
  \textbf{trajectories}\\[1mm]
  $\mathcal T_i=\{u_i^{(0)},u_i^{(1)},\ldots,u_i^\star\}$
};
\draw[flow] (queries) -- (trajectories);

\node[box,draw=solutioncolor,text width=2.55cm] (states) at (8.7+1.2,4.35-0.3) {
  \textcolor{solutioncolor}{\textbf{Converged states}}\\[1mm]
  $U^\star=[u_1^\star\;\cdots\;u_N^\star]$
};

\node[box,draw=correctioncolor,text width=2.55cm] (increments) at (8.7+1.2,2.45-0.3) {
  \textcolor{correctioncolor}{\textbf{Newton}}\\[0mm]
  \textcolor{correctioncolor}{\textbf{increments}}\\[1.2mm]
  $d_i^{(k)}\propto u_i^{(k+1)}-u_i^{(k)}$
};

\draw[blueflow] (trajectories.east) -- ++(3mm,0) |- (states.west);
\draw[orangeflow] (trajectories.east) -- ++(3mm,0) |- (increments.west);

\node[box,draw=solutioncolor,text=neutral,text width=2.75cm]
  (solutionspace) at (12.3+1.2,4.35-0.3) {
  \textcolor{solutioncolor}{\textbf{Solution feature}}\\[0mm]
  \textcolor{solutioncolor}{\textbf{space}}\\[1mm]
  POD basis $\Psi_K$\\[0mm]
  \textit{where solutions lie}
};

\node[box,draw=correctioncolor,text=neutral,text width=2.75cm]
  (correctionspace) at (12.3+1.2,2.45-0.3) {
  \textcolor{correctioncolor}{\textbf{Corrective feature}}\\[0mm]
  \textcolor{correctioncolor}{\textbf{space}}\\[1mm]
  POD basis $\Phi_r$\\[0mm]
  \textit{how Newton moves}
};

\draw[blueflow] (states) -- (solutionspace);
\draw[orangeflow] (increments) -- (correctionspace);

% ---------------------------------------------------------------------------
% ONLINE: two-stage initialization followed by the high-fidelity solve
% ---------------------------------------------------------------------------
\draw[divider] (-0.75,1.15) -- (16.70,1.15);
\node[head,anchor=west] (online) at (-0.7,0.70)
  {\textcolor{neutral}{Online two-stage initialization}};
\node[acc,anchor=west] at (-0.7,0.25)
  {Combining the learned features to produce the Newton initial guess};

\node[box,text width=1.45cm] (newquery) at (0.25,-1.4) {
  \textbf{New query}\\[-0.4mm]
  \textbf{parameter} $\boldsymbol\mu$
};

\node[box,draw=solutioncolor,text width=3.05cm] (prediction) at (3.35,-1.4) {
  \textcolor{solutioncolor}{\textbf{1. Surrogate prediction}}\\[1mm]
  $\widehat u=\bar u+\Psi_K\widehat a(\boldsymbol\mu)$\\[0.2mm]
  \textit{data-driven}\\[-0.4mm]
  \textit{initial estimate}
};

\node[box,draw=correctioncolor,text width=4.8cm] (correction) at (8.05,-1.4) {
  \textcolor{correctioncolor}{
  \textbf{2. Iterative residual-based POD--GMRES correction}
}\\[1mm]

$F_n \approx J(u^{(n)},\boldsymbol\mu)\Phi_r$
\\[-2mm]
$$\beta_n
=
\arg\min_{\beta\in\mathbb R^r}
\left\|
F\!\left(u^{(n)},\boldsymbol\mu\right)+F_n\beta
\right\|_2^2$$

$u^{(n+1)}=u^{(n)}+\Phi_r\beta_n$
};

\node[box,draw=neutral,text width=2.9cm] (newton) at (12.65,-1.4) {
  \textcolor{neutral}{\textbf{High-fidelity Newton solve}}\\[1mm]
  initialized with $u^{(N_{\mathrm{corr}})}$\\[-0.4mm]
};

\node[box,draw=neutral,text=neutral,text width=1.75cm] (solution) at (15.65,-1.4) {
  \textbf{Converged}\\[0mm]
  \textbf{solution}\\[1mm]
  $u^\star(\boldsymbol\mu)$
};

\draw[blueflow] (newquery) -- (prediction);
\draw[orangeflow] (prediction) -- (correction);
\draw[neutralflow] (correction) -- (newton);
\draw[neutralflow] (newton) -- (solution);

\end{tikzpicture}

}

\caption{
Overview of the proposed acceleration strategy.
Offline, the full Newton trajectories generated for the training problems provide two complementary sources of information:
the converged states define the solution feature space $\Psi_K$, whereas the intermediate Newton increments define the corrective feature space $\Phi_r$.
Online, for an unseen parameter $\boldsymbol\mu$, the solution features first provide a surrogate prediction $\widehat u$. Starting from this prediction, successive POD--GMRES iterations minimize the linearized residual over the fixed corrective feature space, with the Jacobian actions approximated by directional finite differences. After $N_{\mathrm{corr}}$ correction iterations, $u^{(N_{\mathrm{corr}})}$ initializes the original high-fidelity Newton solver, which completes convergence to $u^\star(\boldsymbol\mu)$ at the prescribed tolerance.
}
\label{fig:method_overview}

\end{figure*}
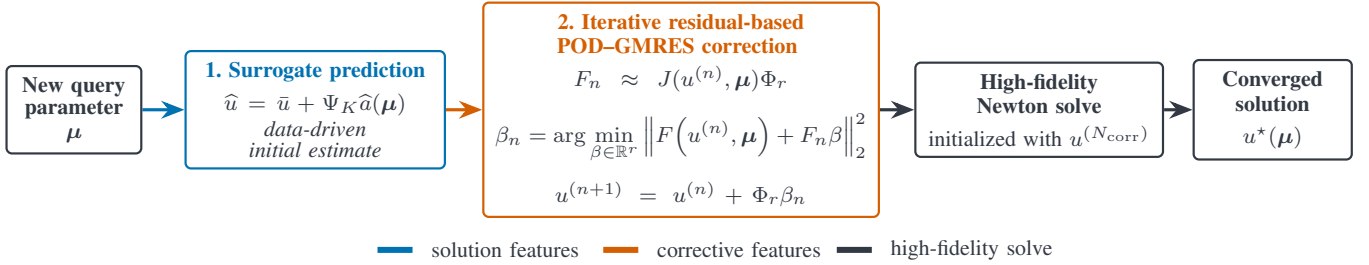

Figure~\ref{fig:method_overview} summarizes the proposed methodology. Two complementary feature spaces are learned from the Newton full trajectory.
The converged states provide solution features for surrogate prediction, whereas their intermediate increments provide corrective features for residual minimization.

\subsection{Problem setting and high-fidelity solver}

The algebraic system obtained after discretizing the nonlinear PDE problem of interest reads in abstract form 
\begin{equation*}
\bm{F}(\bm{u},\bm{\mu})=\bm{0}, \qquad \bm{F}:\R^{n}\times\mathcal{P}\to\R^{n}
\end{equation*}
where $\bm{u}\in\R^{n}$ is the discrete solution, $\bm{\mu}\in\mathcal{P}$ the parameter vector, and $\bm{F}$ the nonlinear residual operator. 
It is assumed to be continuously differentiable with respect to $\bm{u}$ in a neighborhood of the solution.
For a given $\bm{\mu}$, the high-fidelity solver relies on standard Newton's iteration scheme.
Starting from a given initial guess $\bm{u}^{(0)}$, it produces a sequence of iterates defined by
\begin{equation}
\left\{
\begin{aligned}
J(\bm{u}^{(k)}_\mu,\bm{\mu})\,\delta\bm{u}^{(k)}_\mu &= -\bm{F}(\bm{u}^{(k)}_\mu,\bm{\mu}), \\[0.5em]
\bm{u}^{(k+1)}_\mu &= \bm{u}^{(k)}_\mu + \delta\bm{u}^{(k)}_\mu
\end{aligned}
\right.
\label{eq:newton}
\end{equation}
where $J=\partial\bm{F}/\partial\bm{u}$ is the Jacobian matrix of $\bm{F}$ w.r.t. $\bm{u}$ and $\delta\bm{u}^{(k)}$ is the Newton increment. 

%In practice, the Jacobian is not inverted explicitly. Instead, 
%to find $\delta\bm{u}^{(k)}$, 
A linear system is solved (either by a direct method or an iterative one) to get $\delta\bm{u}^{(k)}$.
The iteration is stopped once the relative residual norm 
$\norm{\bm{F}(\bm{u}^{(k)}_\mu,\bm{\mu})}_2/\norm{\bm{F}(\bm{u}^{(0)},\bm{\mu})}_2$ satisfies a given convergence criterion~$\varepsilon_{\mathrm{Newton}}$ being sufficiently small. The last iterate solution is considered the converged solution $\bm{u}_\mu^\star=\bm{u}^\star(\bm{\mu})$.

\subsection{Learning solution features for surrogate prediction}

The upper offline branch of Fig.~\ref{fig:method_overview} corresponds to the learning of solution features. 

There are multiple possible choices of surrogate models for the initial guess.
The surrogate chosen in this work is proper orthogonal decomposition combined with Gaussian process regression (POD-GP).
The proposed methodology is not dependent on this particular selection: the only requirement is the availability of a surrogate capable of providing a more accurate initial estimate than the initialization without a priori knowledge. 

This choice is motivated by the limited-data regime considered in this work.

% Training the surrogate model requires only the parameters and their associated converged solutions. 

Let
$\{(\boldsymbol\mu^i,\boldsymbol u_i^\star)\}_{i=1}^{N_{\mathrm{train}}}$
denote the training dataset, where
$\boldsymbol u_i^\star=\boldsymbol u^\star(\boldsymbol\mu^i)$ is the
converged high-fidelity solution associated with
$\boldsymbol\mu^i$. 

Proper Orthogonal Decomposition is applied to
the centered converged solutions. Retaining the first $K$ POD modes
defines the reduced basis
$\Psi_K=[\boldsymbol\psi_1,\ldots,\boldsymbol\psi_K]$, whose columns
are solution features.
The truncation rank $K$ is defined by the user, or computed automatically according to a 
relative-information-contents (RIC) criterion: 
\[
\frac{\displaystyle{\sum_{\ell=1}^K \sigma_\ell}}{\displaystyle{\sum_{\ell=1}^{N_{\text{train}}} \sigma_\ell}}> 1 - \varepsilon.
\]

The corresponding POD coefficients are obtained by projecting the
training solutions onto $\Psi_K$. Gaussian Process regressors are
then used to learn the parameter-to-coefficient map
$\boldsymbol\mu\mapsto\boldsymbol a(\boldsymbol\mu)$. For an unseen
parameter $\boldsymbol\mu$, the POD--GP prediction is
$$
\widehat{\boldsymbol u}(\boldsymbol\mu)
=
\overline{\boldsymbol u}
+
\Psi_K\,\widehat{\boldsymbol a}(\boldsymbol\mu)
$$
where $\overline{\boldsymbol u}$ is the empirical mean of the
training solutions and
$\widehat{\boldsymbol a}(\boldsymbol\mu)\in\mathbb R^K$ denotes the
coefficients predicted by the Gaussian process regressors. This
prediction constitutes the first stage of the online initialization
strategy and initializes the residual-based POD--GMRES
correction.

In the remainder of the paper,
$\widehat{\boldsymbol u}(\boldsymbol\mu)$ is regarded as the output
of a generic black-box predictor, since the proposed correction does
not depend on the internal construction of the surrogate.

The intermediate Newton iterates produced while solving each training problem are usually unused, although they constitute a valuable source of information.
For a given $\bm{\mu}^i$, the trajectory
$\{\bm{u}^{(0)}_i,\bm{u}^{(1)}_i,\dots, \bm{u}^{(k)}_i,\dots, \bm{u}^\star_i \}$ produced by the Newton solver records the sequence of corrective steps that minimize the residual from the initial guess down to the convergence criterion. 

Standard surrogate construction retains only the converged states and discards the intermediate Newton iterates. However, the increments between consecutive iterates encode directions that were effectively followed by the high-fidelity solver while approaching the solution. 
The lower offline branch of Fig.~\ref{fig:method_overview} exploits this additional information to learn a corrective feature space.

\subsection{Learning corrective search direction features from Newton increments}
%-------------------------------------------------------------------------------
%
Let $\bm{u}^{(k)}_\mu$ and $\bm{u}^{(k+1)}_\mu$ denote two consecutive iterates from a discrete
training trajectory $\{\bm{u}^{(0)},\bm{u}^{(1)}_\mu,\dots, \bm{u}^{(k)}_\mu,\dots, \bm{u}^\star_\mu \}$ for a given sample $\bm{\mu}$.
A unit direction vector at iteration $k$ is computed as
\[
\bm{d}^{(k)}_\mu = \frac{\delta\bm{u}^{(k)}_\mu}{\norm{\delta\bm{u}^{(k)}_\mu}_2}=\frac{\bm{u}^{(k+1)}_\mu-\bm{u}^{(k)}_\mu}{\norm{\bm{u}^{(k+1)}_\mu-\bm{u}^{(k)}_\mu}}_2.
\]
The directions are filtered against a threshold $\varepsilon$:
$$
\frac{\norm{\delta\bm{u}^{(k)}_\mu}_2}{\norm{\delta\bm{u}^{(0)}_\mu}_2}>\varepsilon.
$$
That avoids numerically degenerate pairs, stagnant or already converged steps for which the norm of the increment vector is of the order of the round-off error.

For all sample parameters $\bm{\mu}_i$, the selected unit-norm directions are concatenated into one snapshot matrix $\bm{S}\in\R^{n\times M}$, where $M$ is the total number of directions.
\begin{equation*}
\bm{S} = \begin{bmatrix}
\bm{d}^{(1)}_{\bm{\mu}_1}\! & \!\dots\! & \!\bm{d}^{(k)}_{\bm{\mu}_1} & \!\bm{d}^{(1)}_{\bm{\mu}_2}\! & \!\dots\! & \!\bm{d}^{(k)}_{\bm{\mu}_2} & \!\dots & \!\bm{d}^{(1)}_{\bm{\mu}_p}\! & \!\dots\! & \!\bm{d}^{(k)}_{\bm{\mu}_p}
\end{bmatrix}.
\end{equation*}

This snapshot matrix is relatively large and may contain redundant directions. Then a truncated Singular Value Decomposition (SVD) is performed over $\bm{S}$.
The SVD gives 
$$\bm{S} = \bm{\Phi} \bm{\Sigma} \bm{V}^T$$
where $\bm{\Sigma}=\text{diag}(\sigma_k)$ is the diagonal matrix of nonnegative singular values. 
A truncation at rank $r$ retains only the first $r$ orthonormal columns of~$\bm{\Phi}$.

This yields the semi-orthogonal matrix
$$\bm{\Phi_r} = 
\begin{bmatrix}
    \bm{\phi_1} &
    \bm{\phi_2} &
    \dots &
    \bm{\phi_r}
\end{bmatrix}
$$

The retained POD modes $\{ \bm{\phi}_j\}_{j=1}^r$ are referred to as corrective features. Their span, $\operatorname{Im}(\bm{\Phi_r})$, defines the learned corrective feature space used during the online residual-based correction.
This basis is computed once and for all during the training step. 

\subsection{Online residual-based combination of corrective features}

The online part of Fig.~\ref{fig:method_overview} combines the two learned feature spaces. The solution features first provide the (parameter-dependent) data-driven prediction
$$
u^{(0)}_\mu = \hat{u}(\bm{\mu})
$$

The corrective features then define the fixed low-dimensional space in which successive residual-minimizing corrections are computed. The correction coefficients are determined at each online iteration.

Let $g$ be a prediction operator or a preconditioner, and let $\bm{u}^{(n)}$ denote the current online iterate, initialized by the surrogate prediction, $\bm{u}^{(0)}=\hat{\bm{u}}(\bm{\mu})$. 
The operator $g$ is introduced to keep the formulation consistent with the general nonlinear GMRES formulation, in which $g$ represents an underlying fixed-point iteration or nonlinear preconditioner at each step. 
In the present work, no additional preconditioning step is applied, and $g=\Id$.

The residual and the directional derivatives are evaluated at the predicted point after applying $g$,  $\bar{\bm{u}}^{(n+1)}=g(\bm{u}^{(n)})$. 

Let $\bm{F_n}$ be the directional finite-difference matrix, assembled column by column,
\begin{equation}
\bm{F_{n}}=\begin{bmatrix} \dots &
\dfrac{\bm{F}(\bar{\bm{u}}^{(n+1)}+\eps_{n}\bm{\phi}_{i})-\bm{F}(\bar{\bm{u}}^
{(n+1)})}{\eps_{n}}
& \dots \end{bmatrix},
\end{equation}
whose columns approximate the Jacobian action $J(\bar{\bm{u}}^{(n+1)})\,\bm{\phi}_{i}$ without ever assembling~$J$. 
The correction coefficients minimize the squared norm of the linearized residual over $\im(\Phi_{r})$,
that is
\begin{equation}
\bm{\beta}_{n}=\arg\min_{\bm{\beta}}
\norm{\bm{F}(\bar{\bm{u}}^{(n+1)}) + \bm{F_{n}}\,\bm{\beta}}_2^2
=-\bm{F_{n}}\pinv\,\bm{F}(\bar{\bm{u}}^{(n+1)}),
\end{equation}
where $\bm{F}_n^\dagger$ denotes the Moore--Penrose pseudoinverse of $\bm{F}_n$.

The update reads
\begin{equation}
\bm{u}^{(n+1)}=\bar{\bm{u}}^{(n+1)}+\sum_{j=1}^{r}(\bm{\beta}_{n})_{j}\,\bm{\phi}_{j}.
\label{eq:update}
\end{equation}

Equation~\eqref{eq:update} is a GMRES-type correction step, with the essential difference that the minimization is performed over the fixed, offline subspace $\im(\Phi_{r})$ rather than over a Krylov subspace built online. Thus, each online step requires only $(r+1)$ residual evaluations and a small least-squares problem, and involves neither a Jacobian assembly nor the solution of a linear system of size~$n$.

After $N_{\text{corr}}$ iterations, the resulting state $u^{(N_{\text{corr}})}$ is used as the initial guess for the unchanged high-fidelity Newton solver.
$N_{\text{corr}}$ is determined either by the prescribed maximum number of  iterations or by stagnation, which is detected when the residual reduction between two consecutive iterations is less than $5\%$, i.e., when $
\norm{\bm F(\bm u^{(n+1)})}_2
\geq
0.95\norm{\bm F(\bm u^{(n)})}_2$.

This correction provides no guarantee of convergence to the global minimum of the residual. In practice, the iteration~\eqref{eq:update} is observed to stall at local minima. This behavior is acceptable for the intended purpose: enhancing the quality of the predicted field by lowering the residual norm. The enhanced iterate is then used to initialize the high-fidelity Newton solver, which converges from this better starting point in fewer expensive iterations. The residual evaluations required to assemble $\bm{F_n}$ are inexpensive compared with the several costly Newton steps that are avoided.

\subsection{On the choice of the corrective feature space}
The basis $\Phi_{r}$ is optimal only in a restricted sense. POD provides the feature space that best represents the normalized Newton increments in the mean-square sense, thanks to the Eckart-Young theorem. Optimality for an unseen parameter $\bm{\mu}$ is not guaranteed, and the ideal correction directions for the new problem may lie partly outside $\im(\Phi_{r})$. It remains, however, the best representation available from the training data, and the online step mitigates this limitation by selecting, within this fixed subspace, the combination of directions that minimizes the linearized residual.

The construction of the snapshot matrix requires careful consideration, as it aggregates two distinct classes of search directions. The first class consists of early directions that guide the iterates toward the basin of attraction but do not significantly reduce the residual norm. The second class comprises the last directions, which belong to the asymptotic convergence regime. However, this set is not well suited to representing the dynamics of the pre-basin phase. 
Establishing a selection criterion based on the residual decay value is therefore insufficient to accurately capture directions of interest when the prediction from the surrogate model is not yet in the basin of attraction.  
Applying a truncated Proper Orthogonal Decomposition (POD) to such a heterogeneous collection of directions may result in the omission of some directions that are critical for accurately capturing the initial approach to the basin of convergence.

This assessment tends to favor retaining more POD modes rather than just the first one. Using only the first mode would capture only the overall trend of convergence decay, without accurately representing the behavior at the specific point being analyzed.

Other ideas like building a local snapshot matrix that represents the Newton trajectory behavior are appealing, as the Newton trajectory snapshot matrix is the best subspace possibly defined for the chosen parameter. In practice, it is much more efficient to stick with a global POD basis over the concatenation of directions from every training solve.

\section{Case studies and settings for numerical experiments}
\label{sec:experiences}

The proposed methodology is first assessed on a one-dimensional nonlinear membrane problem serving as a proof-of-concept. It is then tested and evaluated on its two-dimensional extension. 
Both problems involve the same nonlinear constitutive law.

\subsection{One-dimensional Duffing-type problem}

Let $\Omega=(0,1)$. The one-dimensional problem consists of finding $u_\mu:\Omega\rightarrow\mathbb{R}$ such that
\begin{equation}
    -\frac{\partial}{\partial x}
    \left[
        \varphi_\mu\left(\frac{\partial u_\mu}{\partial x}\right)
    \right]
    =q_0
    \quad \text{in } \Omega,
    \label{eq:1d_continuous_problem}
\end{equation}
with homogeneous Dirichlet boundary conditions $u(0) = u(1) = 0$,
where
\begin{equation*}
    \varphi_\mu(s) = \kappa s + \nu s^\gamma
    \label{eq:constitutive_law}
\end{equation*}
defines the constitutive law and $q_0>0$ is a spatially uniform source term.
Regarding the constitutive law, $\kappa>0$ controls the linear contribution, $\nu>0$ controls
the nonlinear part, and $\gamma>1$ is the power that controls the rate of nonlinearity. 
A power law with $\gamma=5$ will be considered for experiments.
The vector $\bm{\mu}=(\kappa,\nu)$ will serve as parameter vector.
The bounded parameter domain $\mathcal{P}\subset\mathbb{R}^2$ used for tests will be detailed later on.

The parameter-independent initial guess used here by the regular Newton solver is
\begin{equation*}
    u^{(0)}(x)
    =
    2\left(
        1-2\,\left| x-\frac{1}{2} \right|
    \right)
    \label{eq:1d_initial_guess}
\end{equation*}

and represented below in Figure~\ref{fig:1d_initial_guess}.

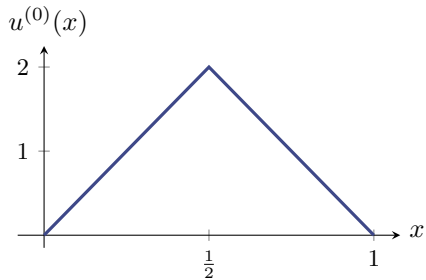
\begin{figure}[ht]
    \centering
    \begin{tikzpicture}
        \begin{axis}[
            width=0.75\columnwidth,
            height=0.48\columnwidth,
            axis lines=middle,
            xmin=-0.08, xmax=1.08,
            ymin=-0.15, ymax=2.25,
            xtick={0,0.5,1},
            xticklabels={$0$,$\frac{1}{2}$,$1$},
            ytick={1,2},
            xlabel={$x$},
            ylabel={$u^{(0)}(x)$},
            xlabel style={
                at={(axis description cs:1,0.08)},
                anchor=west
            },
            ylabel style={
                at={(axis description cs:0.08,1)},
                anchor=south
            },
            tick label style={font=\small},
        ]
            \addplot[
                viridisteal,
                very thick,
                domain=0:1,
                samples=101
            ]
            {2*(1-abs(2*(x-0.5)))};
        \end{axis}
    \end{tikzpicture}
    \caption{Parameter-independent initial guess $u^{(0)}(x)$
    over $\Omega=(0,1)$.}
    \label{fig:1d_initial_guess}
\end{figure}

\subsection{Two-dimensional nonlinear problem}

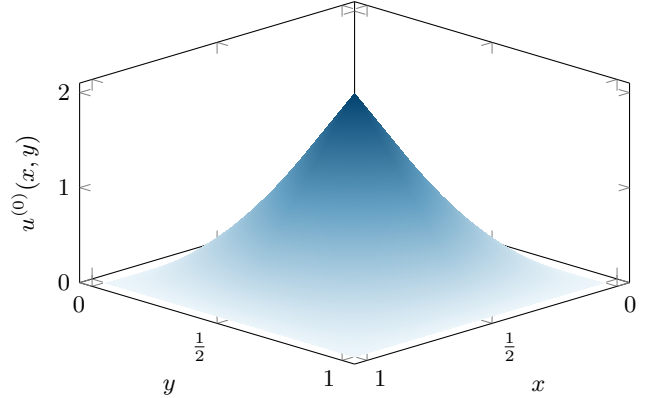
\begin{figure}[ht]
    \centering
    \begin{tikzpicture}
        \begin{axis}[
            width=\columnwidth,
            height=0.72\columnwidth,
            view={135}{30},
            xmin=-0.05, xmax=1.05,
            ymin=-0.05, ymax=1.05,
            zmin=0, zmax=2.1,
            xtick={0,0.5,1},
            xticklabels={$0$,$\frac{1}{2}$,$1$},
            ytick={0,0.5,1},
            yticklabels={$0$,$\frac{1}{2}$,$1$},
            ztick={0,1,2},
            xlabel={$x$},
            ylabel={$y$},
            zlabel={$u^{(0)}(x,y)$},
            tick label style={font=\small},
            label style={font=\small},
            colormap={monoblue}{
                rgb255(0cm)=(235,245,250);
                rgb255(1cm)=(100,160,195);
                rgb255(2cm)=(0,65,110)
            },
        ]
            \addplot3[
                surf,
                shader=interp,
                domain=0:1,
                y domain=0:1,
                samples=41,
                samples y=41,
            ]
            {
                2
                *(1-abs(2*(x-0.5)))
                *(1-abs(2*(y-0.5)))
            };
           
        \end{axis}
    \end{tikzpicture}
    \caption{Parameter-independent initial guess
    $u^{(0)}(x,y)$ over $\Omega=(0,1)^2$.}
    \label{fig:2d_initial_guess}
\end{figure}

Let $\Omega=(0,1)^2$. The two-dimensional problem consists of finding
$u_\mu:\Omega\rightarrow\mathbb{R}$ such that
\begin{equation}
    - \nabla\cdot\boldsymbol{\sigma}(\nabla u_\mu)
    =q_0
    \qquad \text{in } \Omega
    \label{eq:2d_continuous_problem}
\end{equation}
with boundary conditions
\begin{equation*}
    u_\mu=0
    \qquad \text{on } \partial\Omega.
    \label{eq:2d_boundary_conditions}
\end{equation*}
The nonlinear flux is defined componentwise by
\begin{equation*}
    \boldsymbol{\sigma}(\nabla u_\mu)
    =
    \begin{pmatrix}
        \varphi_\mu((u_\mu)_x)\\
        \varphi_\mu((u_\mu)_y)
    \end{pmatrix},
    \qquad
    \varphi_\mu(s) = \kappa s + \nu s^\gamma
    \label{eq:2d_nonlinear_flux}
\end{equation*}
and the parameter vector is still $\bm{\mu}=(\kappa,\nu)$.
The reference Newton solver starts from the initial guess 
\begin{equation*}
    u^{(0)}(x,y)
    =
    2
    \left(1-2\,\left|x-\frac{1}{2}\right|\right)
    \left(1-2\,\left|y-\frac{1}{2}\right|\right).
    \label{eq:2d_initial_guess}
\end{equation*}
Its graphical representation is given in Figure~\ref{fig:2d_initial_guess}.

\subsection{Spatial discretization and numerical implementation}

Both problems are discretized using a conservative flux-difference scheme on uniform grids.

In one dimension, $N_{\mathrm{1D}}=999$ interior nodes are introduced, with grid spacing $h_{\mathrm{1D}}=1/(N_{\mathrm{1D}}+1)=10^{-3}$, resulting in a nonlinear system with $n=999$ degrees of freedom. 
In two dimensions, an $N_{\mathrm{2D}}\times N_{\mathrm{2D}}$ Cartesian grid with $N_{\mathrm{2D}}=50$ interior nodes in each spatial direction is used. The corresponding grid creates a nonlinear system with $n=N_{\mathrm{2D}}^2=2500$ degrees of freedom. 

In both cases, homogeneous Dirichlet boundary conditions are imposed by setting the solution to zero at the boundary nodes.

The residual operators and their Jacobian matrices are implemented using the \texttt{python} library \texttt{JAX}. The Jacobian of the discrete nonlinear residual is obtained by automatic differentiation. At each Newton iteration, the linear system
\begin{equation*}
    \boldsymbol{J}\!\!\left(\boldsymbol{u}_\mu^{(k)}, \bm{\mu}\right)\,
    \boldsymbol{\delta}_\mu^{(k)}
    =
    -\boldsymbol{F}\left(\boldsymbol{u}^{(k)},\bm{\mu}\right)
\end{equation*}
is solved using a direct dense linear solver.
Convergence is stated when the relative residual norm falls below a fixed tolerance~$\varepsilon_{\mathrm{Newton}}$. For both problems, $\varepsilon_{\mathrm{Newton}}$ is set at $10^{-7}$.
\begin{equation*}
    \left\|
        \boldsymbol{F}\left(\boldsymbol{u}^{(k)},\bm{\mu}\right)
    \right\|_2 / \left\|\boldsymbol{F}\left(\boldsymbol{u}^{(0)}\right)
    \right\|_2
    <
    \varepsilon_{\mathrm{Newton}},
\end{equation*}
If the maximum number of Newton iterations is reached, the problem will be considered not converged.

Although the discrete Jacobian matrices possess a sparse structure, all experiments reported in this work currently rely on dense linear algebra. This setting is used to assess the proposed initialization and correction strategy on controlled proof-of-concept problems. 

\subsection{Offline dataset generation}

For both the one- and two-dimensional nonlinear Duffing problems, the variable parameters considered are $\kappa$ and $\nu$. The sampling for training parameters is performed using a Latin Hypercube Sampling technique in the parametric domain. Test parameters are sampled over a uniform grid within the parametric domain. Each parameter is sampled in a min--max range.
Both parameters range from $0.1$ to $10$, such that the parametric domain is defined as
\begin{equation}
    \mathcal{P}
    =
    [\kappa_{\min},\kappa_{\max}]
    \times
    [\nu_{\min},\nu_{\max}]
    =
    [0.1,10]\times[0.1,10].
\end{equation}

\begin{figure}[ht]
    \centering
    \includegraphics[width=\columnwidth]
    {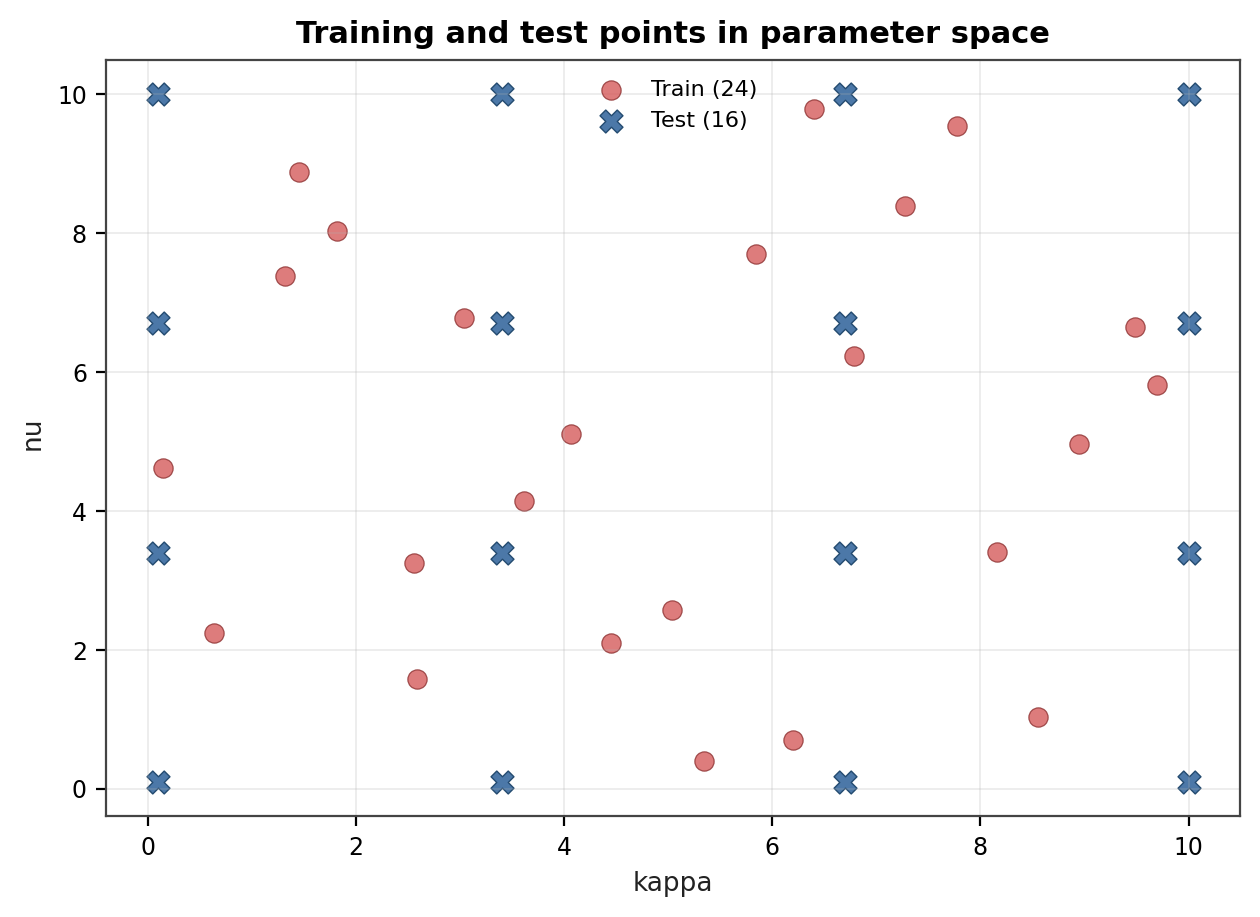}
    \caption{Distribution of the 24 training and 16 test parameter pairs
    over the one-dimensional benchmark parameter domain.}
    \label{fig:parameter_distribution_1d}
\end{figure}

The source amplitude $q_0$ is kept fixed. The dataset contains $N_{\mathrm{train}}=24$ training parameter pairs and $N_{\mathrm{test}}=16$ test parameter pairs. The two sets are disjoint, and their distributions over the parameter domain are shown in Fig.~\ref{fig:parameter_distribution_1d} for the one-dimensional setup. The two-dimensional setup follows the same sampling, and the figure is similar.
With only 24 high-fidelity training solutions available, the study is deliberately set in a small-data regime, reflecting a practical industrial setting in which generating additional nonlinear solutions is computationally expensive.

% --------------------------------------------------
% SECTION VI: RESULTS & DISCUSSION
% --------------------------------------------------
\section{Results \& Discussion}
\label{sec:results}
\subsection{One-dimensional results}
A summary of the online performance for the one-dimensional case is reported in Table~\ref{tab:1d_online_performance}. 
The CPU times in the POD--GMRES contribution include the contributions from the POD--GP prediction initial guess, the POD--GMRES correction step, and the remaining Newton iterations required to reach the prescribed convergence tolerance. 
The speedup is defined as the ratio between the computational time of the cold-start Newton method with default initialization and the total time of the proposed full pipeline.

\begin{table}[!ht]
    \centering
    \caption{Mean online CPU time and speedup for the one-dimensional problem.}
    \label{tab:1d_online_performance}
    \renewcommand{\arraystretch}{1.1}
    \setlength{\tabcolsep}{5pt}
    \begin{tabular}{lccc}
        \hline
        Method & Rank $r$ & Mean CPU (ms) ($\downarrow$) & Speedup ($\uparrow$) \\
        \hline
        Cold Newton & -- & 5590.820 & -- \\
        POD--GP & -- & 1916.016 & $2.92\times$ \\
        POD--GMRES ($10^{-2}$) & 10 & 1360.352 & $4.11\times$ \\
        POD--GMRES ($10^{-4}$) & 23 & 771.484 & $7.25\times$ \\
        POD--GMRES ($10^{-6}$) & 31 & 674.805 & $8.29\times$ \\
        POD--GMRES ($10^{-8}$) & 37 & \textbf{637.280} & $\textbf{8.77}\times$ \\
        \hline
    \end{tabular}
\end{table}

POD--GP initialization alone provides a significant speedup. This speedup remains limited due to the regression error as mentioned. 
The POD--GMRES correction provides a further improvement for any truncation considered. As the truncation rank increases, the improvement increases jointly, meaning a more informative corrective feature space improves the residual decay in the minimization process. 

\subsubsection{Structure of the corrective feature space}

\begin{figure}[!ht]
    \centering
    \includegraphics[width=\columnwidth]{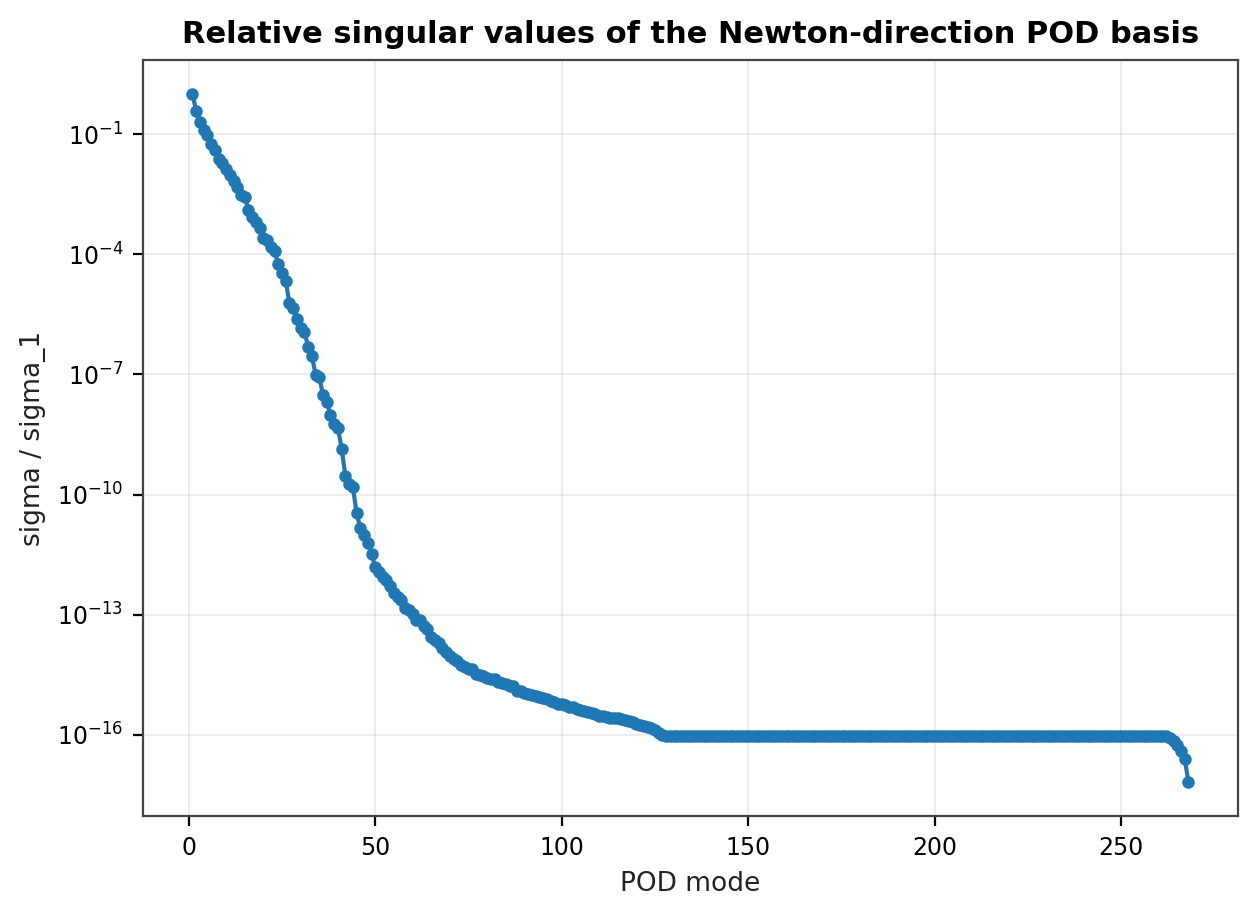}
    \caption{Relative singular values of the corrective snapshot matrix for the one-dimensional problem. The SVD truncation thresholds $10^{-2}$, $10^{-4}$, $10^{-6}$, and $10^{-8}$ retain $10$, $23$, $31$, and $37$ corrective features, respectively.}
    \label{fig:1d_corrective_spectrum}
\end{figure}

Figure~\ref{fig:1d_corrective_spectrum} shows the relative singular value decay of the corrective feature snapshot matrix. A rapid decay is observed and indicates a redundancy among the training trajectories that has been learned by the singular value decomposition.
Four SVD truncation thresholds are considered to construct the basis for the iterative residual-based correction: $10^{-2}$, $10^{-4}$, $10^{-6}$, and $10^{-8}$, which retain $r=10$, $23$, $31$, and $37$ corrective features, respectively.

One important thing to note is that singular-value truncation alone does not determine the best online truncation rank, as it also impacts the size of the least-squares problem to solve. The SVD decay proves optimality in the mean-square reconstruction of the training directions. Online, the directions of importance are those that reduce the residual at unseen parameter values.

\subsubsection{Influence of the SVD truncation on residual reduction}

\begin{figure}[!ht]
    \centering
    \includegraphics[width=\columnwidth]{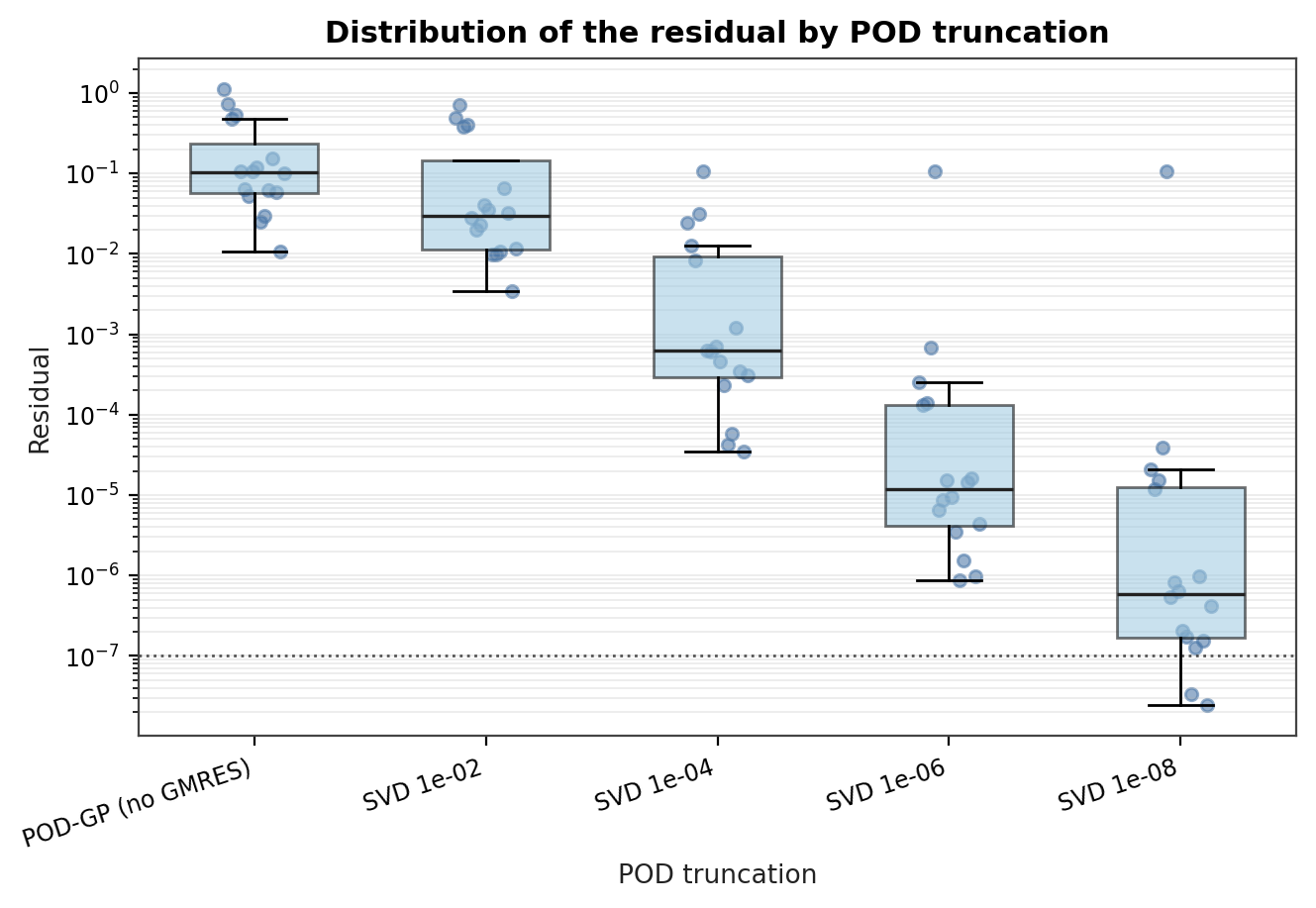}
    \caption{Residual-norm distribution over the 16 one-dimensional test parameters for the POD--GP prediction and after POD--GMRES correction at four SVD truncation thresholds. Boxes show the quartiles and median, individual markers show all test cases, and the dotted horizontal line denotes the prescribed tolerance. Increasing the corrective rank lowers the residual, with two predictions reaching the tolerance directly for $r=37$.}
    \label{fig:1d_residual_by_rank}
\end{figure}

As different truncation ranks have been considered, a truncation effect has been observed. 
Figure~\ref{fig:1d_residual_by_rank} precisely reports this. The impact of the truncation rank applies to the residual norm after the online residual-based correction.
The POD-GP (no GMRES) boxplot reports the residual values of the POD-GP predictions over the test case before applying the online correction.
Each subsequent boxplot applies a GMRES-like correction with different SVD truncation ranks. Retaining a small number of corrective features only provides moderate improvement, whereas increasing the truncation rank shifts the residual downward by several orders of magnitude.

In this one-dimensional case, for two parameter points in the test set, the residual value is observed to fall below the convergence tolerance fixed at $10^{-7}$.

\begin{figure}[!ht]
    \centering
    \includegraphics[width=\columnwidth]{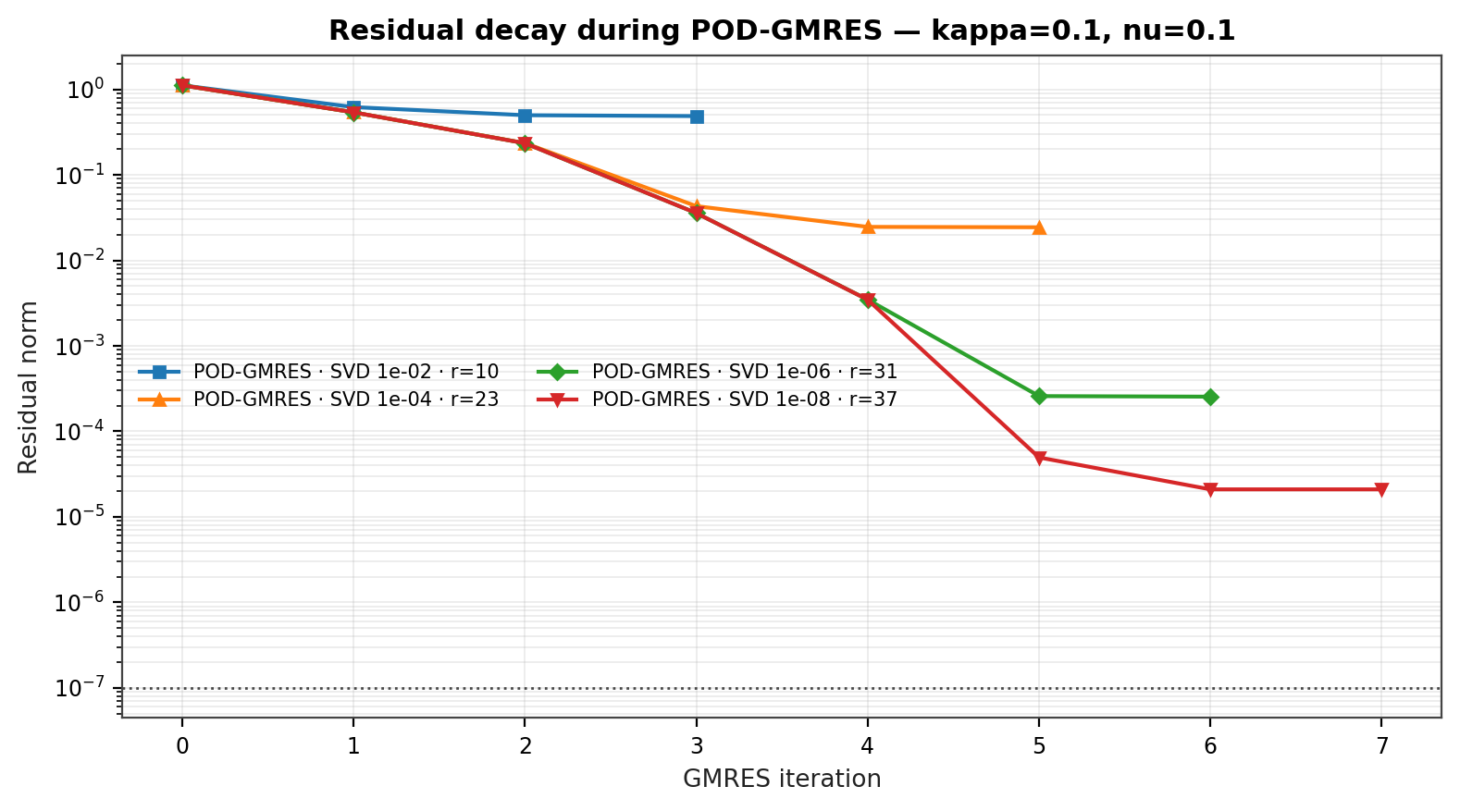}
    \caption{Residual decay during POD--GMRES for the representative one-dimensional test parameter $(\kappa,\nu)=(0.1,0.1)$. Each curve corresponds to an SVD truncation threshold and its associated corrective rank. Larger corrective spaces delay stagnation and reach lower residual plateaus before the Newton restart.}
    \label{fig:1d_gmres_history}
\end{figure}

This improvement of the residual versus the truncation rank is explained in Figure~\ref{fig:1d_gmres_history}. The figure shows the residual decay for one test parameter $(\kappa,\nu)=(0.1,0.1)$ for different truncation ranks over the number of iterations before stagnation.
All configurations initially follow similar trajectories, but the smaller corrective spaces rapidly stagnate as they cannot represent further useful corrections. They only represent the principal corrective features. Increasing the number of retained modes lowers this plateau and allows reaching lower residuals. 
Larger corrective spaces retain more detailed corrective features, which are useful for the residual decay when reaching smaller residual values.

\subsubsection{Impact on the remaining Newton iterations}

\begin{figure}[!ht]
    \centering
    \includegraphics[width=\columnwidth]{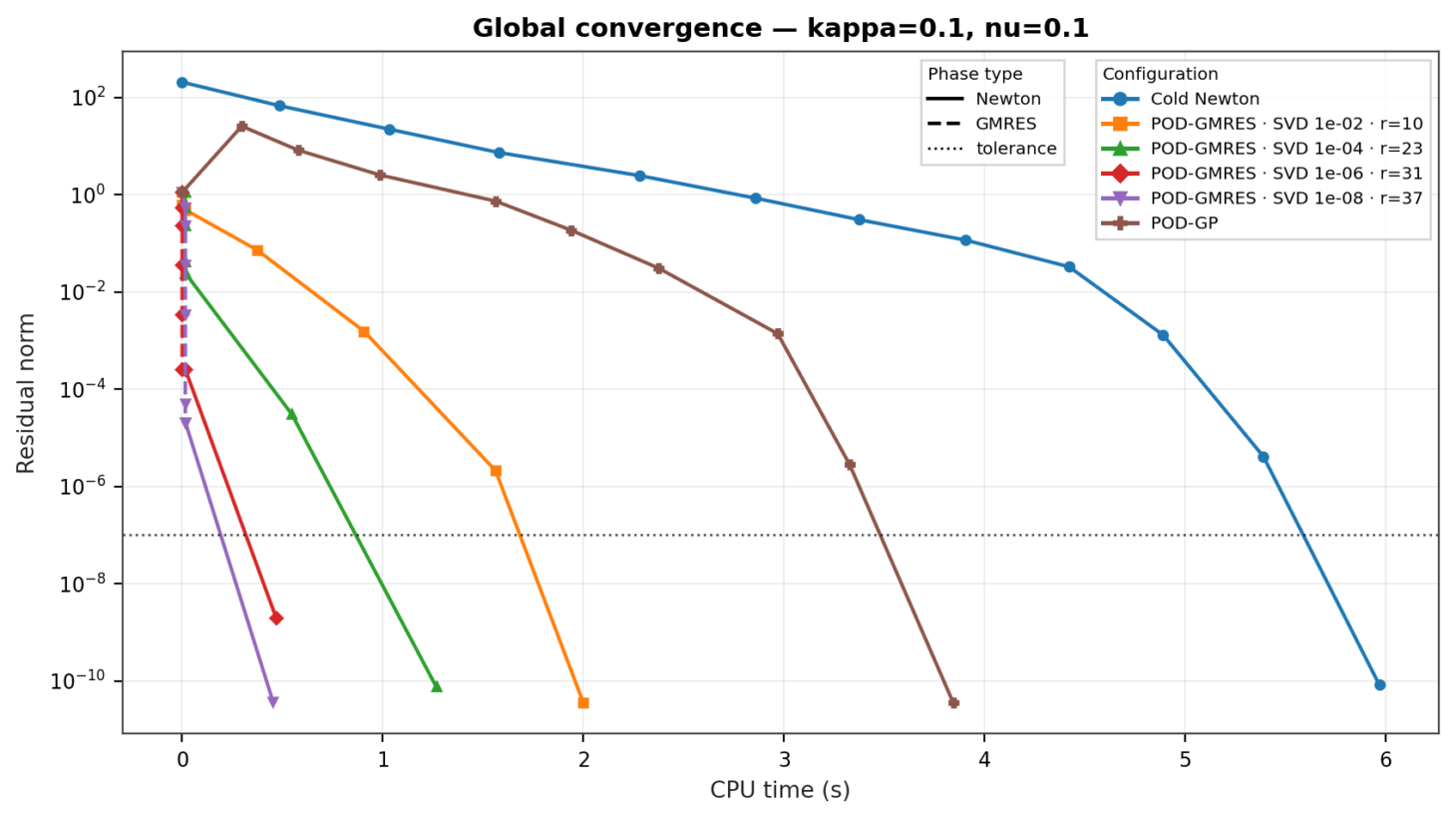}
    \caption{Complete residual histories versus measured online CPU time for the representative one-dimensional test parameter $(\kappa,\nu)=(0.1,0.1)$. Dashed segments denote POD--GMRES corrections, solid segments denote Newton iterations, and the dotted horizontal line denotes the prescribed tolerance.}
    \label{fig:1d_global_convergence}
\end{figure}

After applying the corrective step, the produced predictions are used as initialization of the high-fidelity Newton.
Figure~\ref{fig:1d_global_convergence} illustrates the Newton restart with each method for the same parameter $(\kappa,\nu)=(0.1,0.1)$. 
The x-axis measures the global CPU time, taking into account the time of POD-GP prediction and GMRES correction.
For this test parameter, while POD-GP initialization already reduces the time to solution, adding POD--GMRES further reduces the number of expensive Newton steps. In the large corrective space configuration, the target accuracy is often reached in $1$ iteration.

\begin{figure}[!ht]
    \centering
    \includegraphics[width=\columnwidth]{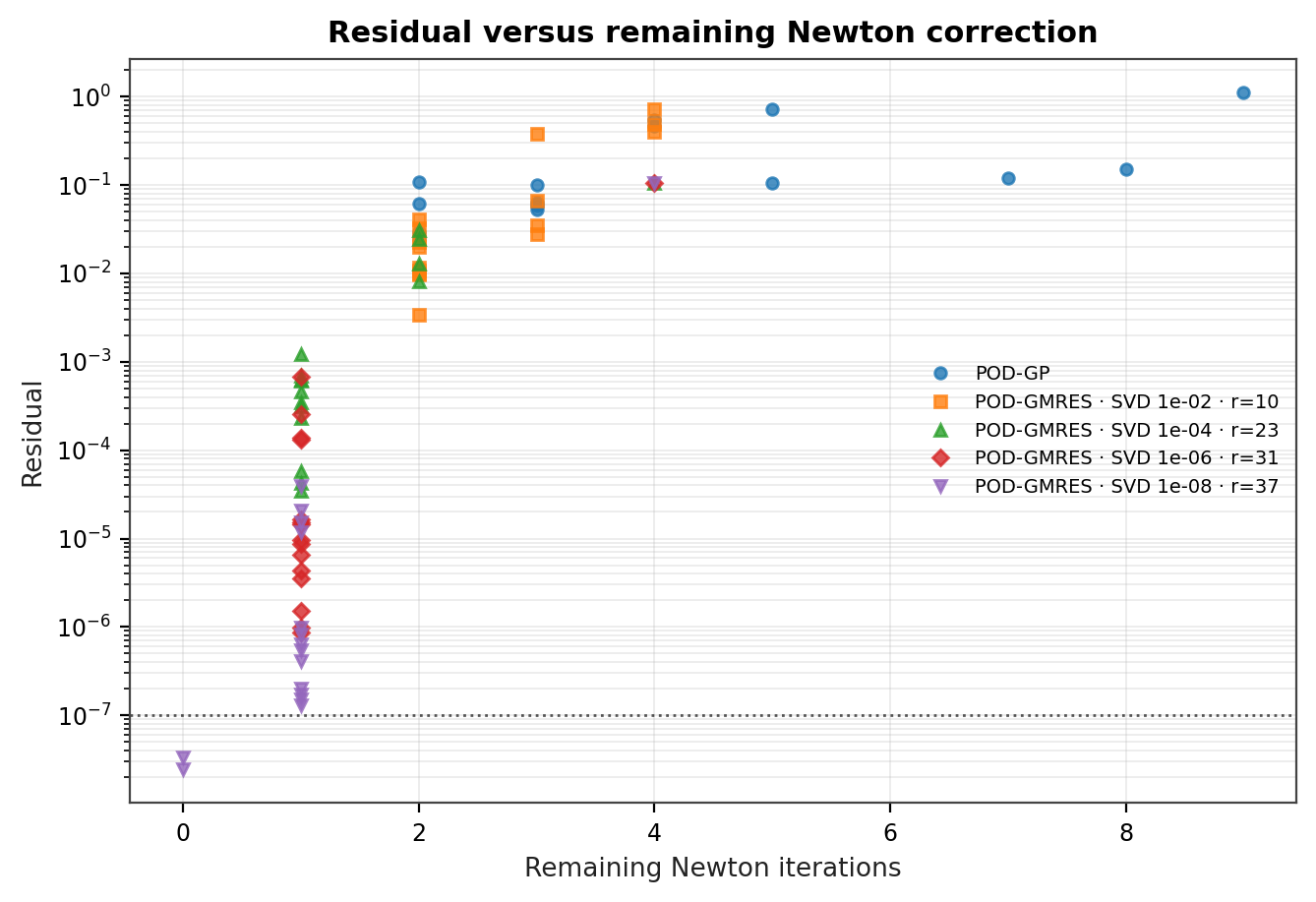}
    \caption{Residual norm after POD--GP initialization or POD--GMRES correction versus the number of remaining Newton iterations for the 16 one-dimensional test parameters. Increasing the corrective rank shifts the points toward lower residuals and shorter Newton phases. Two predictions obtained with $r=37$ already satisfy the tolerance and require no Newton iteration.}
    \label{fig:1d_residual_vs_newton}
\end{figure}

What is notable in Figure~\ref{fig:1d_global_convergence} is that initializing with a lower residual leads to fewer iterations to reach convergence.
This is what Figure~\ref{fig:1d_residual_vs_newton} shows for the whole test set. It connects the initial residual value to the number of remaining Newton iterations. In respect to \citet{jin2024}, reducing the initial residual reduces the number of Newton iterations.
Large POD--GP residuals require several additional Newton iterations, whereas corrected states with residuals below approximately $10^{-3}$ require at most one iteration in nearly all observed cases. Most importantly, two predictions obtained with the $r=37$ corrective space lie directly below the prescribed tolerance and therefore require no subsequent Newton iteration. Newton remains necessary for the other configurations to preserve the same final accuracy.

\begin{figure*}[h!]
    \centering
    \includegraphics[width=0.92\textwidth]{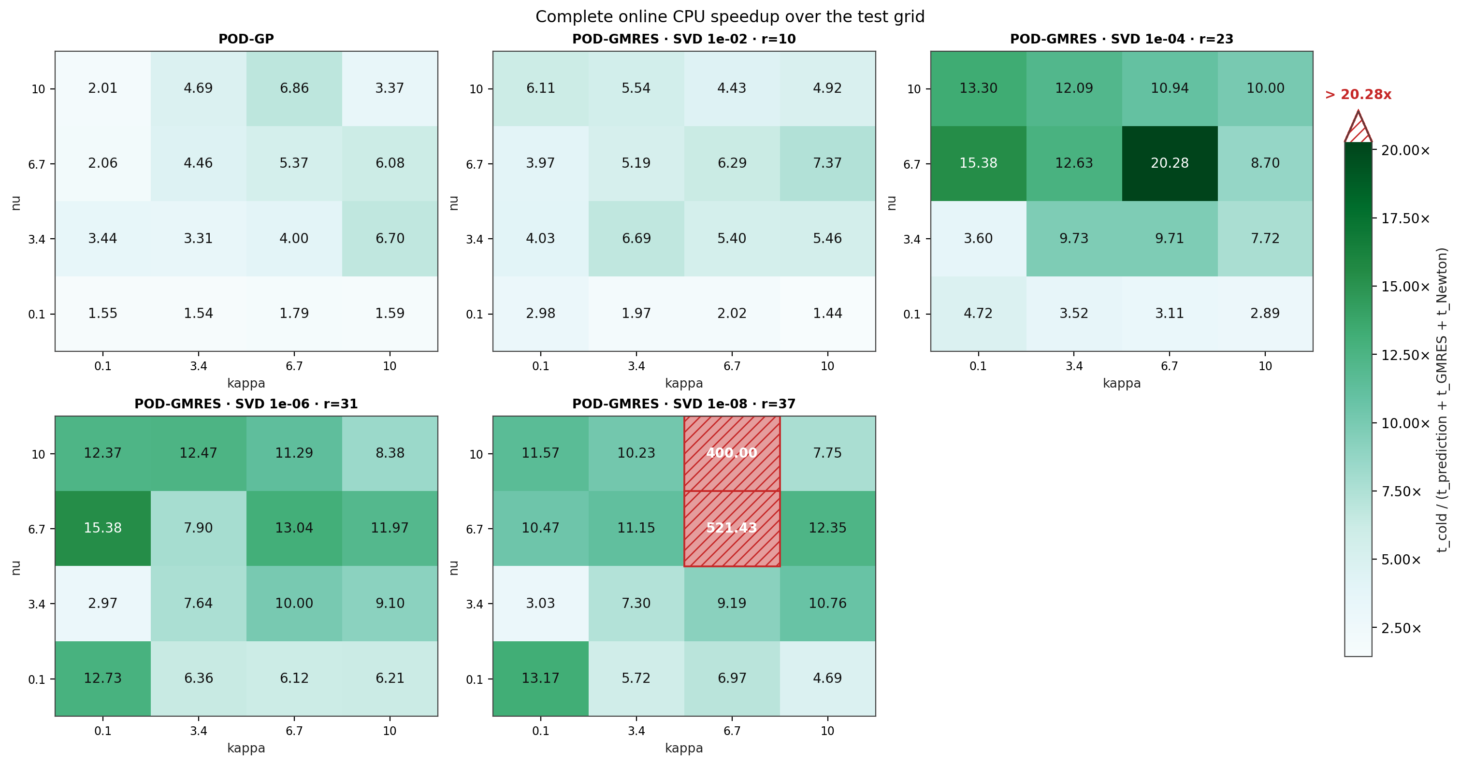}
    \caption{Pointwise speedup of the complete online pipeline relative to cold Newton over the one-dimensional test parameter grid. Each panel corresponds to one initialization configuration. The reported time includes the POD--GP prediction, POD--GMRES when applied, and the remaining Newton iterations required to reach the prescribed tolerance. The two red cells for $r=37$ correspond to direct convergence without Newton and exceed the displayed color scale.}
    \label{fig:1d_speedup_map}
\end{figure*}

Finally, Figure~\ref{fig:1d_speedup_map} reports the pointwise speedup of the complete online procedure relative to cold Newton. POD--GP alone accelerates every test configuration, with speedups ranging from approximately $1.5$ to $6.9$. POD--GMRES provides larger gains when the corrective space is larger. The benefit remains parameter-dependent because both the initial surrogate error and the cost of the Newton iterations avoided by the correction vary with $(\kappa,\nu)$.

The exceptionally large pointwise ratios of $400$ and $521.43$ for $r=37$ occur where the corrected prediction is already converged for the tolerance considered. 
The measured online time becomes very small as not a single Newton iteration is performed. These values demonstrate the possibility of eliminating the Newton phase for particular configurations but should not be interpreted as representative speedups, as most of the time Newton is mandatory after the corrective step.

\subsection{Two-dimensional results}

The two-dimensional setup allows determining whether the learned corrective features remain effective when the spatial dimension and the size of the discrete nonlinear system increase. 

Table~\ref{tab:2d_online_performance} first summarizes the complete online performance. As in the one-dimensional setup, the reported CPU time includes the POD--GP prediction, the POD--GMRES correction when applied, and the remaining Newton iterations.

\begin{table}[!ht]
    \centering
    \caption{Mean online CPU time and speedup for the two-dimensional problem.}
    \label{tab:2d_online_performance}
    \renewcommand{\arraystretch}{1.1}
    \setlength{\tabcolsep}{5pt}
    \begin{tabular}{lccc}
        \hline
        Method & Rank $r$ & Mean CPU (ms) ($\downarrow$) & Speedup ($\uparrow$) \\
        \hline
        Cold Newton & -- & 27589.844 & -- \\
        POD--GP & -- & 10916.016 & $2.53\times$ \\
        POD--GMRES ($10^{-2}$) & 15 & 8793.945 & $3.14\times$ \\
        POD--GMRES ($10^{-4}$) & 53 & 7041.016 & $3.92\times$ \\
        POD--GMRES ($10^{-6}$) & 103 & 4889.648 & $5.64\times$ \\
        POD--GMRES ($10^{-8}$) & 156 & \textbf{3595.703} & $\textbf{7.67}\times$ \\
        \hline
    \end{tabular}
\end{table}

As for the one-dimensional problem, POD--GP initialization alone provides a significant speedup. It reduces the mean online time from $\approx 27$ s to $\approx 10$ s, corresponding to a speedup of $2.53\times$. 

The residual-based correction provides an additional acceleration for every truncation considered. 
The largest corrective space is approximately three times faster than POD--GP initialization alone. This first result already indicates that, despite their additional online cost, the modes retained at lower truncation thresholds remain useful for reducing the complete time to solution.

\subsubsection{Structure of the corrective feature space}

\begin{figure}[H]
    \centering
    \includegraphics[width=\columnwidth]{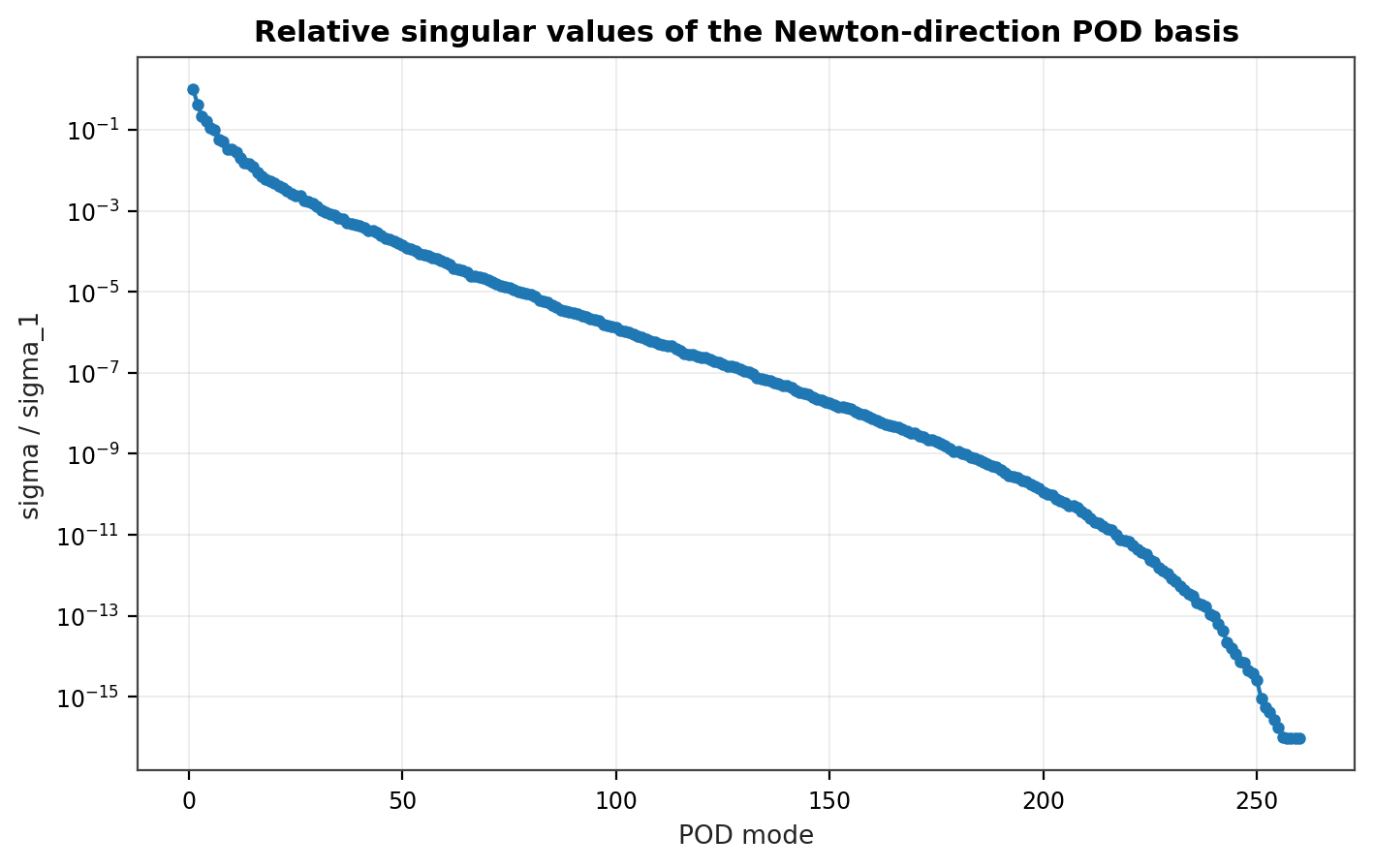}
    \caption{Relative singular values of the corrective snapshot matrix for the two-dimensional problem. The SVD truncation thresholds $10^{-2}$, $10^{-4}$, $10^{-6}$, and $10^{-8}$ retain $15$, $53$, $103$, and $156$ corrective features, respectively.}
    \label{fig:2d_corrective_spectrum}
\end{figure}

Figure~\ref{fig:2d_corrective_spectrum} shows the relative singular values obtained from the Newton-direction snapshots. Compared with the one-dimensional problem, the decay is slower and extends over a larger number of modes. The truncation thresholds $10^{-2}$, $10^{-4}$, $10^{-6}$, and $10^{-8}$ retain $r=15$, $53$, $103$, and $156$ corrective features, respectively. 
The two-dimensional Newton trajectories feature space is more complex to represent and to learn, requiring more modes to reach the same truncation threshold. This indicates a larger variability in the corrective directions.

The slower singular-value decay indicates that the two-dimensional corrective directions are less compressible within a low-dimensional linear subspace. 
At the threshold $10^{-8}$, the retained rank increases from $37$ in one dimension to $156$ in two dimensions. Also, the dimension of one direction is larger as the degree of freedom is larger.
A larger number of modes is required to represent the variability observed among the training directions.

The dimension of the corrective feature space has to be selected wisely as this also impacts the size of the least-squares minimization problem.

\subsubsection{Influence of the SVD truncation on residual reduction}

\begin{figure}[!ht]
    \centering
    \includegraphics[width=\columnwidth]{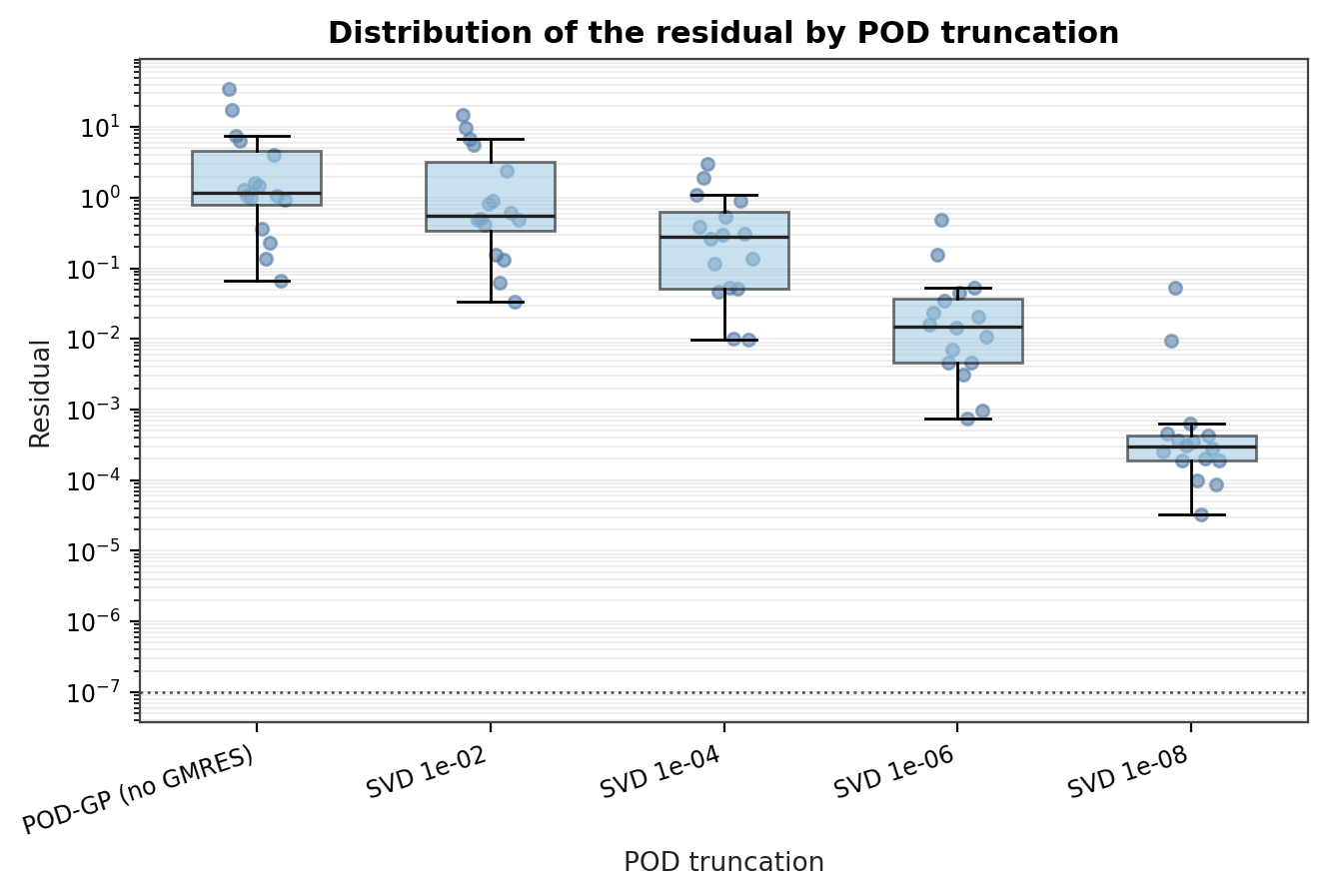}
    \caption{Residual-norm distribution over the 16 two-dimensional test parameters for the POD--GP prediction and after POD--GMRES correction at four SVD truncation thresholds. Boxes show the quartiles and median, individual markers show all test cases, and the dotted horizontal line denotes the prescribed tolerance. Increasing the corrective rank progressively lowers the residual.}
    \label{fig:2d_residual_by_rank}
\end{figure}

Figure~\ref{fig:2d_residual_by_rank} reports the residual-norm distributions over the 16 test parameters before and after the POD--GMRES correction. The POD--GP predictions initially produce residual norms mostly between $10^{-1}$ and $10^{1}$. The correction with $r=15$ provides only a moderate reduction, whereas the distribution progressively shifts toward lower residual levels as additional modes are retained. With an SVD truncation threshold of $10^{-8}$, the corrected residual norms are concentrated approximately between $10^{-5}$ and $10^{-3}$. 

Despite the lower compressibility of the two-dimensional directions, the additional modes remain useful during the online correction. 
Retaining higher-order modes progressively improves the correction. 

However, this reduction requires a substantial increase in the corrective rank, up to $r=156$, reflecting the slower singular-value decay and the lower compressibility of the two-dimensional corrective directions.
Even with this larger space, the correction remains less effective than in one dimension, where residual levels of approximately $10^{-6}$ to $10^{-7}$ were reached.

\subsubsection{Impact on the remaining Newton iterations}

\begin{figure}[!ht]
    \centering
    \includegraphics[width=\columnwidth]{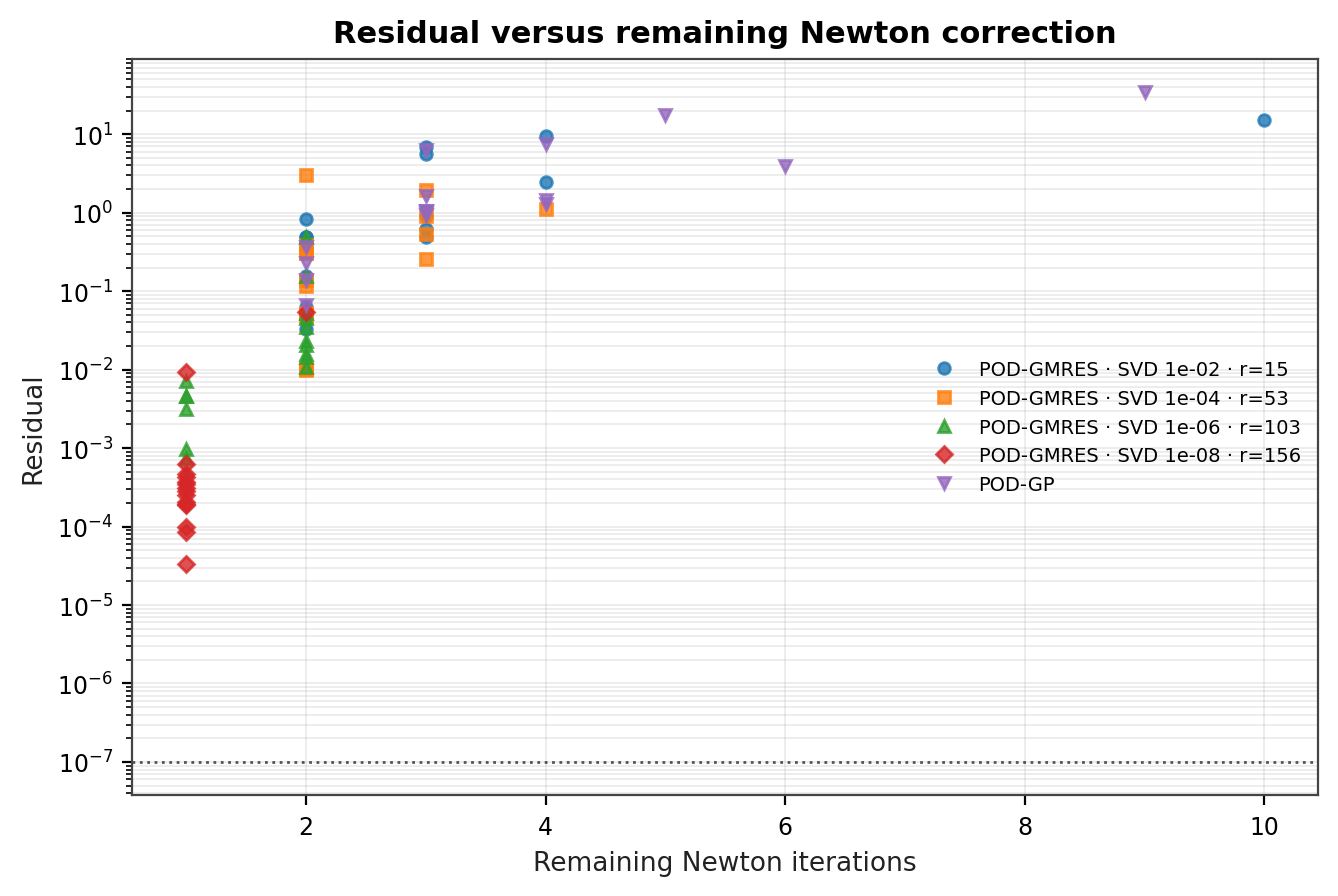}
    \caption{Residual norm after POD--GP initialization or POD--GMRES correction versus the number of remaining Newton iterations for the 16 two-dimensional test parameters. Increasing the corrective rank shifts the points toward lower residuals and shorter Newton phases. All predictions obtained with $r=156$ require only one Newton iteration.}
    \label{fig:2d_residual_vs_newton}
\end{figure}

The relationship between the corrected residual and the number of remaining Newton iterations is shown in Fig.~\ref{fig:2d_residual_vs_newton}.

The same trend as in the one-dimensional case is observed: although the residual norm does not uniquely determine the remaining Newton work, it provides an upper bound on the number of iterations. Reducing the residual before the Newton restart shortens the final Newton phase.

Unlike the one-dimensional experiment, none of the corrected two-dimensional predictions directly reaches the prescribed tolerance. A final Newton step therefore remains necessary to guarantee the same accuracy for every test configuration.

\begin{figure}[!ht]
    \centering
    \includegraphics[width=\columnwidth]{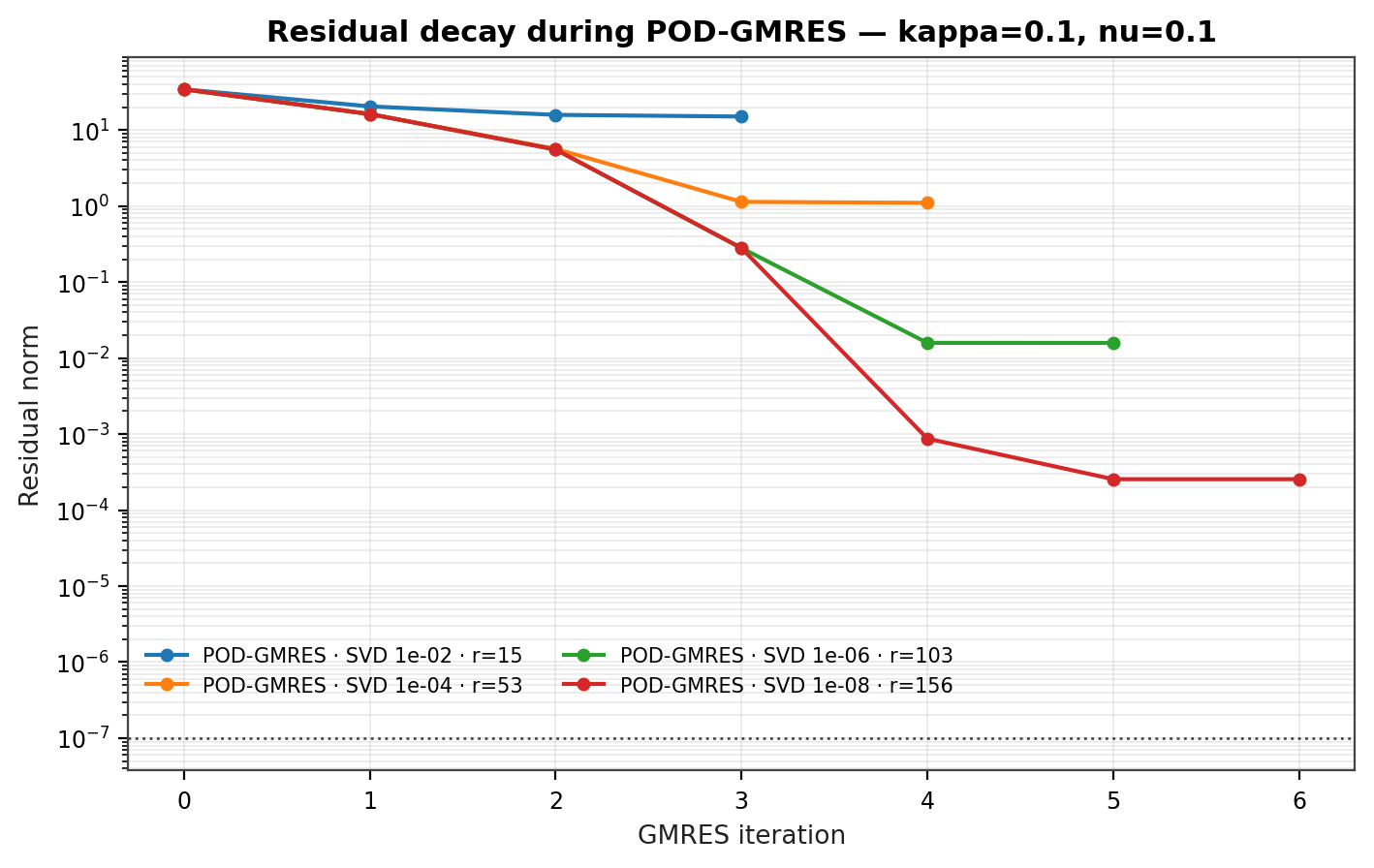}
    \caption{Residual decay during POD--GMRES for the representative two-dimensional test parameter $(\kappa,\nu)=(0.1,0.1)$. Each curve corresponds to an SVD truncation threshold and its associated corrective rank. Larger corrective spaces delay stagnation and reach lower residual plateaus before the Newton restart.}
    \label{fig:2d_podgmres_decay}
\end{figure}

Figure~\ref{fig:2d_podgmres_decay} reports the residual decay during the POD--GMRES correction for the representative parameter configuration $(\kappa,\nu)=(0.1,0.1)$.

The same behavior as in the one-dimensional case is observed: all corrective spaces initially follow a similar residual-decay trajectory, but the smaller spaces rapidly stagnate once their available correction directions have been exhausted. 
Accordingly, increasing the corrective rank preserves this initial decay while allowing POD--GMRES to continue minimizing the residual, with the largest space reaching approximately $10^{-4}$.

\begin{figure}[!ht]
    \centering
    \includegraphics[width=\columnwidth]{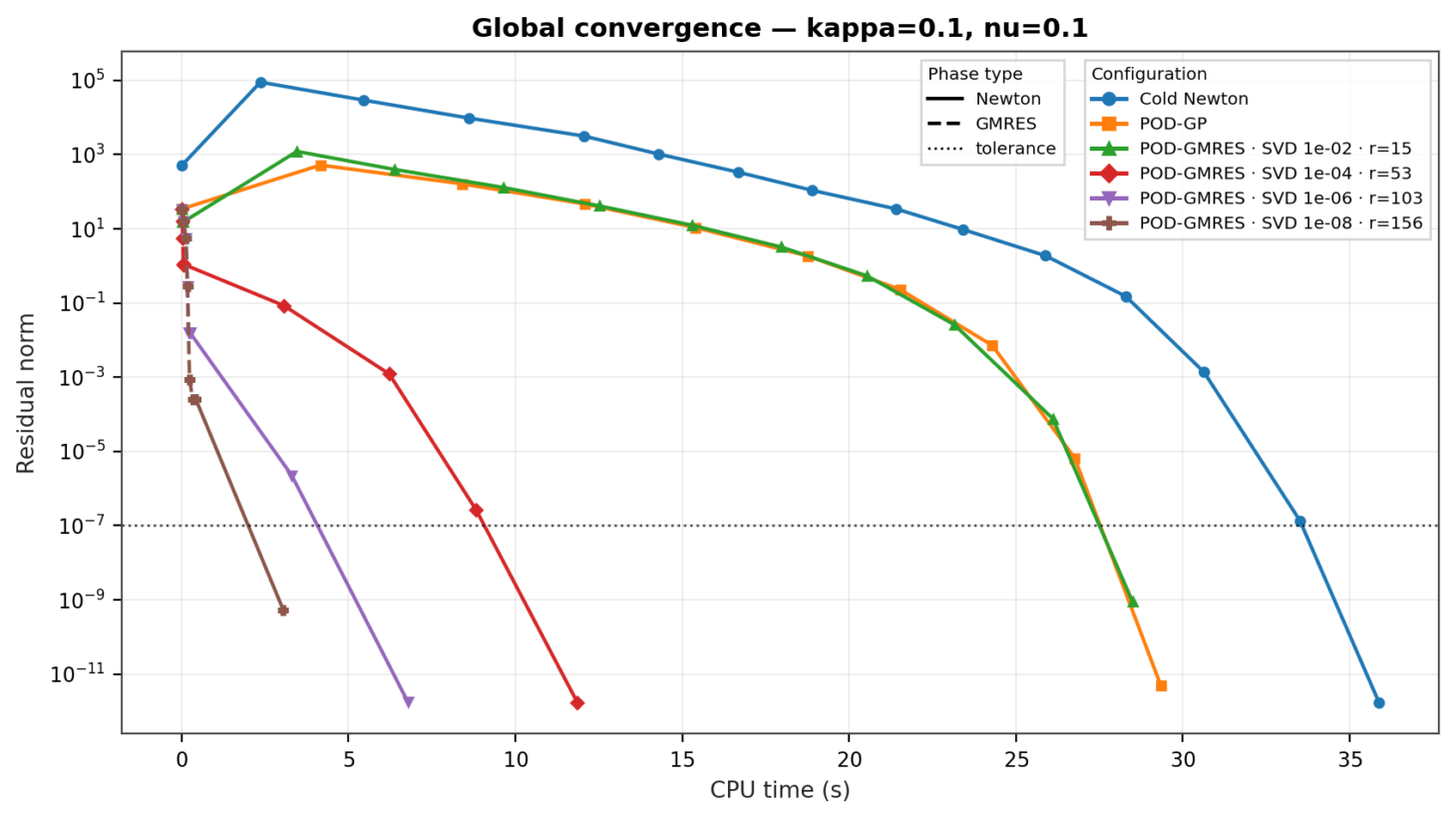}
    \caption{Complete residual histories versus measured online CPU time for the representative two-dimensional test parameter $(\kappa,\nu)=(0.1,0.1)$. Dashed segments denote POD--GMRES corrections, solid segments denote Newton iterations, and the dotted horizontal line denotes the prescribed tolerance.}
    \label{fig:2d_global_convergence}
\end{figure}

Figure~\ref{fig:2d_global_convergence} shows the corresponding convergence histories for the complete online pipeline.

Cold Newton requires approximately $36$ s for this configuration, while POD--GP initialization reduces the measured online CPU time to approximately $29$ s. The two smallest corrective spaces provide little additional benefit because their initial residual is approximately the same as that of POD-- GP. The benefit becomes more important for the larger truncation ranks $r=103$ and $r=156$, for which the lower residual at the end of POD--GMRES considerably shortens the remaining Newton solve.

\begin{figure*}[!ht]
    \centering
    \includegraphics[width=\textwidth]{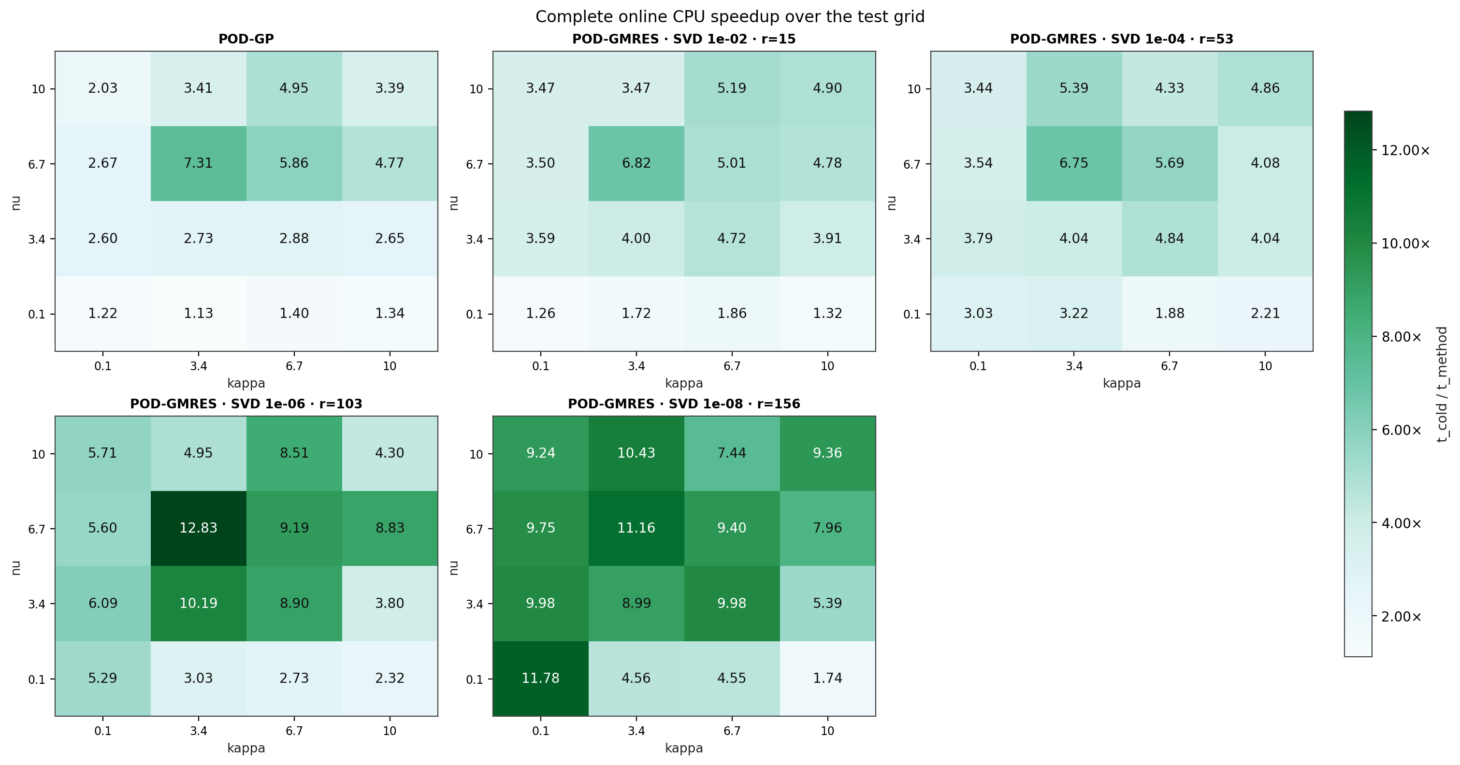}
    \caption{Pointwise speedup of the complete online pipeline relative to cold Newton over the two-dimensional test parameter grid. Each panel corresponds to one initialization configuration. The reported time includes the POD--GP prediction, POD--GMRES when applied, and the remaining Newton iterations required to reach the prescribed tolerance.}
    \label{fig:2d_speedup_heatmap}
\end{figure*}

Figure~\ref{fig:2d_speedup_heatmap} reports the pointwise speedup of the complete online pipeline over the test parameter grid. POD--GP improves the computational time for every test configuration, although its benefit remains moderate and parameter dependent.

Although the pointwise speedup is not strictly monotonic with the corrective rank, the largest corrective space yields the highest mean speedup.

Overall, the two-dimensional experiment confirms that the learned corrective features remain transferable to unseen parameters, but it also reveals a limitation of the POD representation. The greater variability among the Newton directions leads to a slower singular-value decay and requires a substantially larger corrective space than in one dimension.

\subsection{Discussion}

The one- and two-dimensional experiments support the same methodology. The solution feature space first provides a parameter-dependent approximation, but the residual of this prediction remains limited by both the POD truncation and the regression error. The corrective feature space then supplies corrective directions learned from the intermediate Newton iterates, allowing the online procedure to reduce the residual beyond the prediction plateau before handing the state to the original Newton solver.

Retaining more corrective modes generally allows POD--GMRES to reach a lower residual before the Newton restart, which in turn reduces the number of remaining Newton iterations.
The remaining Newton count is not a single-valued function of the residual norm because the local nonlinear behavior also depends on the parameter value. Nevertheless, the residual is a clear indicator of the amount of nonlinear work left, and the overall trend remains valid across the test sets. Thus, reducing the residual before the Newton restart benefits the subsequent Newton solve.

The proposed methodology produces comparable relative gains in both benchmarks. POD--GP alone provides mean speedups of $2.92\times$ in one dimension and $2.53\times$ in two dimensions. With the largest corrective spaces, these values increase to $8.77\times$ and $7.67\times$, respectively.

Since all configurations are ultimately assessed using the same Newton stopping criterion, this reduction in computational time does not result from a relaxation of the prescribed final accuracy, thereby ensuring a fair comparison. Thanks to this final Newton stage, the computed solutions preserve the accuracy and robustness of the high-fidelity Newton solves.

\section{Concluding remarks}
\label{sec:conclusion}
%===========================
In this work, we proposed a two-stage initialization strategy for accelerating the solution of nonlinear parametrized PDE problems by learning complementary features from previous Newton solutions. Rather than considering Newton's method solely as a nonlinear solver, the proposed approach exploits the information contained in complete Newton trajectories. Two reduced spaces are extracted offline: a solution feature space, used to predict an initial approximation, and a corrective search direction feature space, learned from intermediate Newton increments and used online to reduce the residual through a Jacobian-free POD–GMRES correction procedure.

A key feature of the proposed methodology is its weakly intrusive character. Unlike projection-based reduced-order methods or nonlinear preconditioning techniques requiring access to the discretization or Jacobian assembly, the proposed correction relies only on residual field evaluations and the solution of a small least-squares problem. The original high-fidelity Newton solver therefore remains completely unchanged and still guarantees the final solution accuracy, while the learned correction acts only as an efficient nonlinear initializer. This makes the methodology particularly attractive for large industrial simulation codes, where modifying the solver internals is often impractical. The price to pay is a larger database that stores not only the converged solution but also the whole Newton iteration path.

The numerical experiments demonstrate that exploiting Newton trajectory information significantly improves standard surrogate-based initialization. Compared with a POD-GP predictor alone, the learned corrective search space consistently decreases the initial residual norm, reduces the number of remaining Newton iterations, and yields substantial reductions in overall computational time.

For the one-dimensional benchmark, speedups approaching one order of magnitude are obtained, while some test cases converge immediately after the correction step, eliminating the need for any subsequent Newton iteration. 
Quite often, the approximate solution returned by the correction step falls into the quadratic convergence region of Newton's method.
More generally, the observed reduction in Newton iterations follows the expected relationship between initialization quality and convergence efficiency, confirming that learning corrective search directions constitutes an effective complement to classical surrogate prediction.

Several directions naturally emerge from this work. The corrective search direction feature space could be enriched through adaptive or locally parameter-dependent bases instead of a single global POD basis (e.g., using Grassmann or Stiefel manifold basis interpolation \cite{stiefel-interpol, zimmermann2022manifoldinterpolationmodelreduction}). In parallel, alternative surrogate models -- including neural operators or foundation models for scientific computing -- could replace the POD–GP predictor without modifying the correction framework.
Extending the methodology to large-scale industrial multiphysics problems solved with Jacobian-Free Newton–Krylov algorithms also appears particularly promising, since the proposed correction naturally fits this computational setting. 
Such an extension would, however, require addressing the treatment of multiple physical fields.
The present experiments involve a single field, whereas, in coupled or multiphysics problems, Newton increments belong to a product space and combine corrections associated with different physical variables, such as velocity and pressure. Constructing a meaningful corrective space would therefore require accounting for the different scales and mathematical roles of these fields while ensuring that the resulting mixed increments remain compatible with the structure and constraints of the coupled problem.

A further perspective concerns the dependence of the learned corrective space on the spatial discretization. In the present formulation, Newton increments are represented as discrete vectors and can therefore be directly reused only on the same mesh. An extension to multiple meshes could consist in learning corrective directions at the functional level and discretizing them on the target mesh only during the online residual-minimization phase. Such a formulation would separate the learned corrective information from a particular discretization and could allow the same directional knowledge to be transferred across meshes.

Finally, a rigorous theoretical analysis of the convergence properties of the nonlinear POD–GMRES correction and the approximation capabilities of the learned corrective space would provide valuable insight into the observed numerical performance.

Beyond the proposed algorithm itself, this work advocates a new way of exploiting nonlinear solvers: learning not only the solution manifold but also the dynamics of the solver that generates it. We believe that exploiting these trajectories as reusable computational knowledge opens a new avenue for weakly intrusive acceleration methods and, more broadly, for data-driven nonlinear solvers.

\vspace{3em}
\phantomsection
\bibliographystyle{unsrtnat}
\bibliography{references}

\begin{thebibliography}{39}
\providecommand{\natexlab}[1]{#1}
\providecommand{\url}[1]{\texttt{#1}}
\expandafter\ifx\csname urlstyle\endcsname\relax
  \providecommand{\doi}[1]{doi: #1}\else
  \providecommand{\doi}{doi: \begingroup \urlstyle{rm}\Url}\fi

\bibitem[Saad and van~der Vorst(2001)]{saad2001iterative}
Yousef Saad and Henk van~der Vorst.
\newblock Iterative solution of linear systems in the 20th century.
\newblock In \emph{Numerical Analysis: Historical Developments in the 20th
  Century}, pages 193--198. Elsevier Science Publishers, 2001.
\newblock ISBN 0-444-50617-9.

\bibitem[van~der Vorst(2003)]{vandervorst2003krylov}
H.~A. van~der Vorst.
\newblock \emph{Iterative Krylov Methods for Large Linear Systems}.
\newblock Cambridge University Press, 2003.
\newblock ISBN 0-521-81828-1.

\bibitem[Knoll and Keyes(2004)]{Knoll_2004}
D.A. Knoll and D.E. Keyes.
\newblock Jacobian-free newton–krylov methods: a survey of approaches and
  applications.
\newblock \emph{Journal of Computational Physics}, 193:\penalty0 357--397,
  2004.
\newblock \doi{10.1016/j.jcp.2003.08.010}.
\newblock URL \url{https://doi.org/10.1016/j.jcp.2003.08.010}.

\bibitem[Allgower and Georg(2011)]{numerical-continuation}
Eugene~L Allgower and Kurt Georg.
\newblock \emph{Numerical Continuation Methods}.
\newblock Springer Series in Computational Mathematics. Springer, Berlin,
  Germany, October 2011.

\bibitem[Jin et~al.(2025)Jin, Maierhofer, Schratz, and Xiang]{jin2024}
Tianyu Jin, Georg Maierhofer, Katharina Schratz, and Yang Xiang.
\newblock A fast neural hybrid newton solver adapted to implicit methods for
  nonlinear dynamics, 2025.
\newblock URL \url{https://arxiv.org/abs/2407.03945}.

\bibitem[McGreivy et~al.(2024)McGreivy, Nick, Hakim, and Ammar]{weakbaselines}
McGreivy, Nick, Hakim, and Ammar.
\newblock Weak baselines and reporting biases lead to overoptimism in machine
  learning for fluid-related partial differential equations.
\newblock \emph{Nature Machine Intelligence}, 6\penalty0 (10):\penalty0
  1256–1269, 2024.
\newblock \doi{10.1038/s42256-024-00897-5}.
\newblock URL \url{http://dx.doi.org/10.1038/s42256-024-00897-5}.

\bibitem[Guo and Hesthaven(2018)]{podgp}
Mengwu Guo and Jan~S. Hesthaven.
\newblock Reduced order modeling for nonlinear structural analysis using
  gaussian process regression.
\newblock \emph{Computer Methods in Applied Mechanics and Engineering},
  341:\penalty0 807--826, 2018.
\newblock ISSN 0045-7825.
\newblock \doi{https://doi.org/10.1016/j.cma.2018.07.017}.
\newblock URL
  \url{https://www.sciencedirect.com/science/article/pii/S0045782518303487}.

\bibitem[Geelen et~al.(2023)Geelen, Wright, and Willcox]{quadraticmanifolds}
Rudy Geelen, Stephen Wright, and Karen Willcox.
\newblock Operator inference for non-intrusive model reduction with quadratic
  manifolds.
\newblock \emph{Computer Methods in Applied Mechanics and Engineering},
  403:\penalty0 115717, 2023.
\newblock \doi{https://doi.org/10.1016/j.cma.2022.115717}.
\newblock URL
  \url{https://www.sciencedirect.com/science/article/pii/S0045782522006727}.

\bibitem[Lu et~al.(2021)Lu, Lu, Jin, Pengzhan, Pang, Guofei, Zhang, Zhongqiang,
  Karniadakis, and Em]{deeponet}
Lu, Lu, Jin, Pengzhan, Pang, Guofei, Zhang, Zhongqiang, Karniadakis, and George
  Em.
\newblock Learning nonlinear operators via deeponet based on the universal
  approximation theorem of operators.
\newblock \emph{Nature Machine Intelligence}, 3\penalty0 (3):\penalty0
  218–229, 2021.
\newblock \doi{10.1038/s42256-021-00302-5}.
\newblock URL \url{http://dx.doi.org/10.1038/s42256-021-00302-5}.

\bibitem[Li et~al.(2021)Li, Kovachki, Azizzadenesheli, Liu, Bhattacharya,
  Stuart, and Anandkumar]{FNO}
Zongyi Li, Nikola Kovachki, Kamyar Azizzadenesheli, Burigede Liu, Kaushik
  Bhattacharya, Andrew Stuart, and Anima Anandkumar.
\newblock Fourier neural operator for parametric partial differential
  equations, 2021.
\newblock URL \url{https://arxiv.org/abs/2010.08895}.

\bibitem[Li et~al.(2020)Li, Kovachki, Azizzadenesheli, Liu, Bhattacharya,
  Stuart, and Anandkumar]{GKN}
Zongyi Li, Nikola Kovachki, Kamyar Azizzadenesheli, Burigede Liu, Kaushik
  Bhattacharya, Andrew Stuart, and Anima Anandkumar.
\newblock Neural operator: Graph kernel network for partial differential
  equations, 2020.
\newblock URL \url{https://arxiv.org/abs/2003.03485}.

\bibitem[Raonić et~al.(2023)Raonić, Molinaro, Ryck, Rohner, Bartolucci,
  Alaifari, Mishra, and de~Bézenac]{raonietal2023}
Bogdan Raonić, Roberto Molinaro, Tim~De Ryck, Tobias Rohner, Francesca
  Bartolucci, Rima Alaifari, Siddhartha Mishra, and Emmanuel de~Bézenac.
\newblock Convolutional neural operators for robust and accurate learning of
  pdes.
\newblock \emph{arXiv preprint}, 2023.
\newblock URL \url{http://arxiv.org/abs/2302.01178v3}.

\bibitem[Taghikhani et~al.(2025)Taghikhani, Yamazaki, Varghese, Apel, Asl, and
  Rezaei]{taghikhanietal2025}
Kianoosh Taghikhani, Yusuke Yamazaki, Jerry~Paul Varghese, Markus Apel,
  Reza~Najian Asl, and Shahed Rezaei.
\newblock Neural-initialized newton: Accelerating nonlinear finite elements via
  operator learning.
\newblock \emph{arXiv preprint}, 2025.
\newblock URL \url{http://arxiv.org/abs/2511.06802v1}.

\bibitem[Kadeethum et~al.(2022)Kadeethum, O’Malley, Ballarin, Ang, Fuhg,
  Bouklas, Silva, Salinas, Heaney, Pain, Lee, Viswanathan, and
  Yoon]{Kadeethum_2022}
Teeratorn Kadeethum, Daniel O’Malley, Francesco Ballarin, Ida Ang, Jan~N.
  Fuhg, Nikolaos Bouklas, Vinicius L.~S. Silva, Pablo Salinas, Claire~E.
  Heaney, Christopher~C. Pain, Sanghyun Lee, Hari~S. Viswanathan, and Hongkyu
  Yoon.
\newblock Enhancing high-fidelity nonlinear solver with reduced order model.
\newblock \emph{Scientific Reports}, 12, 2022.
\newblock \doi{10.1038/s41598-022-22407-6}.
\newblock URL \url{https://doi.org/10.1038/s41598-022-22407-6}.

\bibitem[Tiba et~al.(2024)Tiba, Azzeddine, Dairay, Thibault, Vuyst, Florian,
  Mortazavi, Iraj, Ramirez, and Juan-Pedro]{tiba2022fsi}
Tiba, Azzeddine, Dairay, Thibault, De~Vuyst, Florian, Mortazavi, Iraj, Berro
  Ramirez, and Juan-Pedro.
\newblock Non-intrusive reduced order models for partitioned fluid–structure
  interactions.
\newblock \emph{Journal of Fluids and Structures}, 128:\penalty0 104156, 2024.
\newblock \doi{10.1016/j.jfluidstructs.2024.104156}.
\newblock URL \url{http://dx.doi.org/10.1016/j.jfluidstructs.2024.104156}.

\bibitem[Bergmann and Iollo(2016)]{bergmann2016ot}
Michel Bergmann and Angelo Iollo.
\newblock Bioinspired swimming simulations.
\newblock \emph{Journal of Computational Physics}, 323:\penalty0 310--321,
  2016.
\newblock \doi{https://doi.org/10.1016/j.jcp.2016.07.022}.
\newblock URL
  \url{https://www.sciencedirect.com/science/article/pii/S0021999116303175}.

\bibitem[Aghili et~al.(2025)Aghili, Franck, Hild, Michel-Dansac, and
  Vigon]{FNO_efranck}
Joubine Aghili, Emmanuel Franck, Romain Hild, Victor Michel-Dansac, and Vincent
  Vigon.
\newblock Accelerating the convergence of newton’s method for nonlinear
  elliptic pdes using fourier neural operators.
\newblock \emph{Communications in Nonlinear Science and Numerical Simulation},
  140:\penalty0 108434, January 2025.
\newblock ISSN 1007-5704.
\newblock \doi{10.1016/j.cnsns.2024.108434}.
\newblock URL \url{http://dx.doi.org/10.1016/j.cnsns.2024.108434}.

\bibitem[Lechevallier et~al.(2025)Lechevallier, Desroziers, Faney, Flauraud,
  and Nataf]{lechevallier}
Antoine Lechevallier, Sylvain Desroziers, Thibault Faney, Eric Flauraud, and
  Fr{\'e}d{\'e}ric Nataf.
\newblock Hybrid newton method for the acceleration of well event handling in
  the simulation of co2 storage using supervised learning.
\newblock \emph{Computers \& Geosciences}, 197:\penalty0 105872, 2025.
\newblock \doi{10.1016/j.cageo.2025.105872}.

\bibitem[Dong et~al.(2020)Dong, Xie, Kestor, and Li]{donti2020smartpgsim}
Wenqian Dong, Zhen Xie, Gokcen Kestor, and Dong Li.
\newblock Smart-pgsim: Using neural network to accelerate ac-opf power grid
  simulation, 2020.
\newblock URL \url{https://arxiv.org/abs/2008.11827}.

\bibitem[Novello et~al.(2022)Novello, Poëtte, Lugato, Peluchon, and
  Congedo]{novello2022nn}
Paul Novello, Gaël Poëtte, David Lugato, Simon Peluchon, and Pietro~Marco
  Congedo.
\newblock Accelerating hypersonic reentry simulations using deep learning-based
  hybridization (with guarantees), 2022.
\newblock URL \url{https://arxiv.org/abs/2209.13434}.

\bibitem[Huang et~al.(2019)Huang, Wang, and Yang]{intdeep}
Jianguo Huang, Haoqin Wang, and Haizhao Yang.
\newblock Int-deep: A deep learning initialized iterative method for nonlinear
  problems.
\newblock \emph{arXiv preprint}, 2019.
\newblock URL \url{http://arxiv.org/abs/1910.01594v6}.

\bibitem[Carlberg et~al.(2013)Carlberg, Farhat, Cortial, and Amsallem]{GNAT}
Kevin Carlberg, Charbel Farhat, Julien Cortial, and David Amsallem.
\newblock The gnat method for nonlinear model reduction: Effective
  implementation and application to computational fluid dynamics and turbulent
  flows.
\newblock \emph{Journal of Computational Physics}, 242:\penalty0 623--647,
  2013.
\newblock \doi{10.1016/j.jcp.2013.02.028}.
\newblock URL \url{https://doi.org/10.1016/j.jcp.2013.02.028}.

\bibitem[Barrault et~al.(2004)Barrault, Maday, Nguyen, and Patera]{EIM}
Maxime Barrault, Yvon Maday, Ngoc~Cuong Nguyen, and Anthony~T. Patera.
\newblock An ‘empirical interpolation’ method: application to efficient
  reduced-basis discretization of partial differential equations.
\newblock \emph{Comptes Rendus. Mathématique}, 339:\penalty0 667--672, 2004.
\newblock \doi{10.1016/j.crma.2004.08.006}.
\newblock URL \url{https://doi.org/10.1016/j.crma.2004.08.006}.

\bibitem[Chaturantabut and Sorensen(2009)]{DEIM}
Saifon Chaturantabut and Danny~C. Sorensen.
\newblock Discrete empirical interpolation for nonlinear model reduction.
\newblock \emph{Proceedings of the 48th IEEE Conference on Decision and Control
  (CDC) held jointly with 2009 28th Chinese Control Conference}, pages
  4316--4321, 2009.
\newblock \doi{10.1109/CDC.2009.5400045}.
\newblock URL \url{https://doi.org/10.1109/cdc.2009.5400045}.

\bibitem[Everson and Sirovich(1995)]{gappyPOD}
R.~Everson and L.~Sirovich.
\newblock Karhunen–loève procedure for gappy data.
\newblock \emph{Journal of the Optical Society of America A}, 12:\penalty0
  1657, 1995.
\newblock \doi{10.1364/JOSAA.12.001657}.
\newblock URL \url{https://doi.org/10.1364/josaa.12.001657}.

\bibitem[Cai and Keyes(2002)]{caikeyes2002aspin}
Xiao-Chuan Cai and David~E. Keyes.
\newblock Nonlinearly preconditioned inexact {N}ewton algorithms.
\newblock \emph{SIAM Journal on Scientific Computing}, 24\penalty0
  (1):\penalty0 183--200, 2002.
\newblock \doi{10.1137/S106482750037620X}.

\bibitem[Dolean et~al.(2016)Dolean, Gander, Kheriji, Kwok, and
  Masson]{dolean2016raspen}
Victorita Dolean, Martin~J. Gander, Walid Kheriji, Felix Kwok, and Roland
  Masson.
\newblock Nonlinear preconditioning: How to use a nonlinear {S}chwarz method to
  precondition {N}ewton's method.
\newblock \emph{SIAM Journal on Scientific Computing}, 38\penalty0
  (6):\penalty0 A3357--A3380, 2016.
\newblock \doi{10.1137/15M102887X}.

\bibitem[Benzi(2002)]{benzi2002preconditioning}
Michele Benzi.
\newblock Preconditioning techniques for large linear systems: A survey.
\newblock \emph{Journal of Computational Physics}, 182\penalty0 (2):\penalty0
  418--477, 2002.
\newblock \doi{10.1006/jcph.2002.7176}.

\bibitem[Chen(2005)]{chen2005preconditioning}
Ke~Chen.
\newblock \emph{Matrix Preconditioning Techniques and Applications}.
\newblock Cambridge University Press, 2005.
\newblock ISBN 978-0521838283.

\bibitem[Axelsson(1996)]{axelsson1996iterative}
Owe Axelsson.
\newblock \emph{Iterative Solution Methods}.
\newblock Cambridge University Press, 1996.
\newblock ISBN 978-0-521-55569-2.

\bibitem[Wang et~al.(2021)Wang, Wang, and
  Perdikaris]{wang2021learningsolutionoperatorparametric}
Sifan Wang, Hanwen Wang, and Paris Perdikaris.
\newblock Learning the solution operator of parametric partial differential
  equations with physics-informed deeponets, 2021.
\newblock URL \url{https://arxiv.org/abs/2103.10974}.

\bibitem[Azizzadenesheli et~al.(2024)Azizzadenesheli, Kovachki, Li,
  Liu-Schiaffini, Kossaifi, and Anandkumar]{no_nvidia}
Kamyar Azizzadenesheli, Nikola Kovachki, Zongyi Li, Miguel Liu-Schiaffini, Jean
  Kossaifi, and Anima Anandkumar.
\newblock Neural operators for accelerating scientific simulations and design,
  2024.
\newblock URL \url{https://arxiv.org/abs/2309.15325}.

\bibitem[Li et~al.(2023)Li, Zheng, Kovachki, Jin, Chen, Liu, Azizzadenesheli,
  and Anandkumar]{li2023physicsinformedneuraloperatorlearning}
Zongyi Li, Hongkai Zheng, Nikola Kovachki, David Jin, Haoxuan Chen, Burigede
  Liu, Kamyar Azizzadenesheli, and Anima Anandkumar.
\newblock Physics-informed neural operator for learning partial differential
  equations, 2023.
\newblock URL \url{https://arxiv.org/abs/2111.03794}.

\bibitem[Cheng et~al.(2026)Cheng, Sahadath, Yang, Pan, and Ji]{cheng2026}
Qiyun Cheng, Md~Hossain Sahadath, Huihua Yang, Shaowu Pan, and Wei Ji.
\newblock Md-pnop: Equation-recast neural operators for minimal-data
  extrapolation and pde solver acceleration, 2026.
\newblock URL \url{https://arxiv.org/abs/2509.01416}.

\bibitem[Lee et~al.(2025)Lee, Youngkyu, Liu, Shanqing, Darbon, Jerome,
  Karniadakis, and Em]{lee2025neural}
Lee, Youngkyu, Liu, Shanqing, Darbon, Jerome, Karniadakis, and George Em.
\newblock A neural-operator preconditioned newton method for accelerated
  nonlinear solvers.
\newblock \emph{arXiv preprint arXiv:2511.08811}, 2025.

\bibitem[Ding and Wang(2025)]{dingwang2025}
Renjie Ding and Dongling Wang.
\newblock Adaptive residual-driven newton solver for nonlinear systems of
  equations.
\newblock \emph{arXiv preprint}, 2025.
\newblock URL \url{http://arxiv.org/abs/2501.03487v1}.

\bibitem[Luo and Cai(2023)]{pinl}
Li~Luo and Xiao-Chuan Cai.
\newblock \(\text{PIN}^{\mathcal l}\) : Preconditioned inexact newton with
  learning capability for nonlinear system of equations.
\newblock \emph{SIAM Journal on Scientific Computing}, 45\penalty0
  (2):\penalty0 A849--A871, 2023.
\newblock \doi{10.1137/22M1507942}.
\newblock URL \url{https://doi.org/10.1137/22M1507942}.

\bibitem[{El Omari} et~al.(2025){El Omari}, {El Khlifi}, and
  Cordier]{stiefel-interpol}
Achraf {El Omari}, Mohamed {El Khlifi}, and Laurent Cordier.
\newblock Stiefel manifold interpolation for non-intrusive model reduction of
  parameterized fluid flow problems.
\newblock \emph{Journal of Computational Physics}, 521:\penalty0 113564, 2025.
\newblock ISSN 0021-9991.
\newblock \doi{https://doi.org/10.1016/j.jcp.2024.113564}.
\newblock URL
  \url{https://www.sciencedirect.com/science/article/pii/S002199912400812X}.

\bibitem[Zimmermann(2022)]{zimmermann2022manifoldinterpolationmodelreduction}
Ralf Zimmermann.
\newblock Manifold interpolation and model reduction, 2022.
\newblock URL \url{https://arxiv.org/abs/1902.06502}.

\end{thebibliography}

\end{document}